\documentclass[11pt]{article}
\usepackage[T1]{fontenc}
\usepackage{bm}
\usepackage{color}
\usepackage{colortbl}
\usepackage{hyperref}
\usepackage{graphicx}
\usepackage[letterpaper,margin=1 in]{geometry}

\usepackage{tabularx}

\usepackage{xcolor}
\usepackage[numbers,sort&compress]{natbib}
 
\usepackage{capt-of}
\usepackage{tabularx}
\usepackage{adjustbox}

\usepackage{pifont}
\usepackage{subcaption}

\definecolor{bestgreen}{RGB}{214,239,214}
\definecolor{secondorange}{RGB}{248,229,215}
\newcommand{\litmark}{$^{\diamond}$}

\usepackage[english]{babel}
\usepackage{xspace}
\usepackage{algorithm,algorithmicx}
\usepackage{tcolorbox}
\usepackage{tikz}

\usepackage{framed}
\usepackage{algpseudocode}
\usepackage{amsthm,nccmath}
\usepackage{amsmath,bm}
\usepackage[normalem]{ulem}
\usepackage{bigints}
\usepackage{amssymb}
\usepackage{caption}
\usepackage{graphicx}

\usepackage{enumerate}
\usepackage{multirow}
\usepackage{booktabs}
\usepackage{hyperref}
\usepackage{url}
\usepackage{color, colortbl}
\usepackage{graphicx}
\usepackage{amsmath,amssymb, bm}
\usepackage{algorithm}
\usepackage{algpseudocode}

\usepackage{flushend}
\usepackage{enumitem}
\usepackage[bottom]{footmisc}
\usepackage{hyperref}
\hypersetup{
	colorlinks   = true, %Colours links instead of ugly boxes
	urlcolor     = blue, %Colour for external hyperlinks
	linkcolor    = blue, %Colour of internal links
	citecolor    = blue   %Colour of citations
}
\newcommand{\algo}{\textsc{\textbf{INFUSE}}}

\usepackage{float}

\usepackage{epstopdf}
\usepackage{footnote}
\makesavenoteenv{tabular}
\usepackage{bbm}
\usepackage{amsthm}
\usepackage[makeroom]{cancel}
\usepackage{stmaryrd}
\usepackage{amsmath,bm}
\usepackage{amssymb}
\usepackage{mathtools, cuted}
\usepackage{kantlipsum,setspace}

\theoremstyle{plain}
\newtheorem{theorem}{Theorem}[section]

\newtheorem*{lem*}{Lemma}

\newtheorem*{cor*}{Corollary}

\newtheorem*{defn*}{Definition}

\newtheorem{proposition}[theorem]{Proposition}
\theoremstyle{definition}
\newtheorem{definition}{Definition}[section]
\newtheorem*{definition*}{Definition}

\theoremstyle{remark}

\newtheorem*{rem*}{Remark}

\let\hat\widehat
\let\tilde\widetilde

\title{{\bf   Hallucination Mitigation for Large Vision-Language Models via Implicit Feature Stabilization}}
\date{} 
 
\author{\large Aditi Sarker$^\dagger$, Rafi Ibn Sultan$^\dagger$, Hui Zhu$^\dagger$, Dongxiao Zhu$^{\dagger, \ast}$, \\
and Prashant Khanduri$^{\dagger, \ast}$  \\[.5cm]
\small $^{\dagger}$Department of Computer Science, Wayne State University, MI, USA  \\
\small $^{\ast}$Institute for AI and Data Science (AIDaS), Wayne State University, MI, USA \\
\small Email:
\texttt{\{aditi1, rafis, hui, dzhu, khanduri.prashant\}@wayne.edu}}

\begin{document}
\maketitle

\begin{abstract}
Large Vision-Language Models (LVLMs) are prone to hallucinations: they fluently describe objects, attributes, and scenes that are not in the image. We connect part of this failure to a measurable property of their representations, \emph{feature instability}, where mild semantics-preserving perturbations of the input cause large changes in the learned embeddings; hallucination rates rise together with this variability. Existing stability-motivated remedies are \emph{explicit}, in the sense that they intervene at inference time through latent steering or constrained decoding, and pay for it on every query. We propose \emph{implicit} stabilization instead: perturbation-invariance is built into the model weights during fine-tuning, and nothing extra runs at deployment. Our framework, \algo, first stabilizes visual and textual representations around perturbation-averaged and ground-truth anchors, then aligns the stabilized representations across modalities with bidirectional contrastive objectives. We prove that the anchor's root-mean-square deviation from the perturbation-mean representation shrinks at rate $1/\sqrt{K}$ in the number of views, and that under a Lipschitz decoder, this bounds how much any perturbation can change the model's hallucination behavior. On LLaVA-1.5, LLaVA-1.6, and Qwen3-VL-8B-Instruct, \algo\ reduces AMBER CHAIR by 46-63\% relative to each base model, improves ObjHal, MMHal, HallusionBench, and POPE, and preserves VQA-v2 and TextVQA, all with no inference-time overhead.
\end{abstract}

\section{Introduction}
\label{sec:intro}

\begin{figure}[t]
    \centering
    \includegraphics[width=0.6\textwidth]{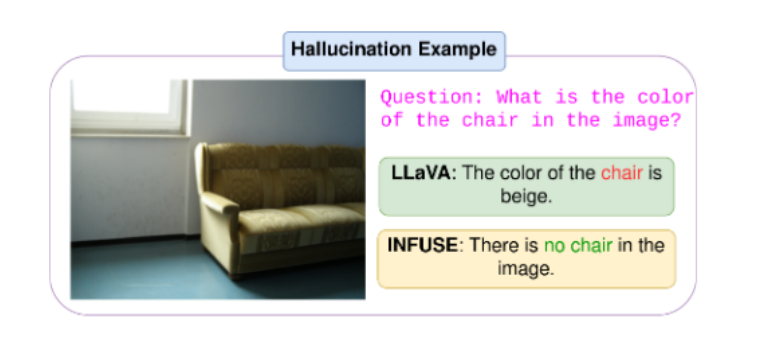}
    \caption{\textbf{Object-existence hallucination.} The LVLM incorrectly describes a non-existent \textcolor{red}{chair} in an image containing only a sofa.}
    \label{fig:objexist}
\end{figure}

Large Vision-Language Models (LVLMs) couple pre-trained vision encoders with large language models and now perform well across captioning, visual question answering (VQA), and instruction following \cite{radford2021learning,li2023blip,liu2024llavanext}, which has pushed them into settings where reliability matters \cite{lee2025multimodality,tu2024towards}. They remain prone to \emph{hallucinations}: fluent, plausible responses unsupported by the visual evidence (Figure \ref{fig:objexist}). The failure mode is well documented on dedicated benchmarks such as POPE, AMBER, MMHal-Bench, and HallusionBench \cite{li2023evaluating,wang2023amber,sun2024aligning,guan2024hallusionbench}, and surveyed at length \cite{liu2024survey,bai2024hallucination,huang2025survey,sahoo2024comprehensive}.
\vspace{2 mm}

\begin{figure}[t]
    \centering
    \begin{subfigure}{\columnwidth}
        \centering
\includegraphics[width=\linewidth,keepaspectratio]{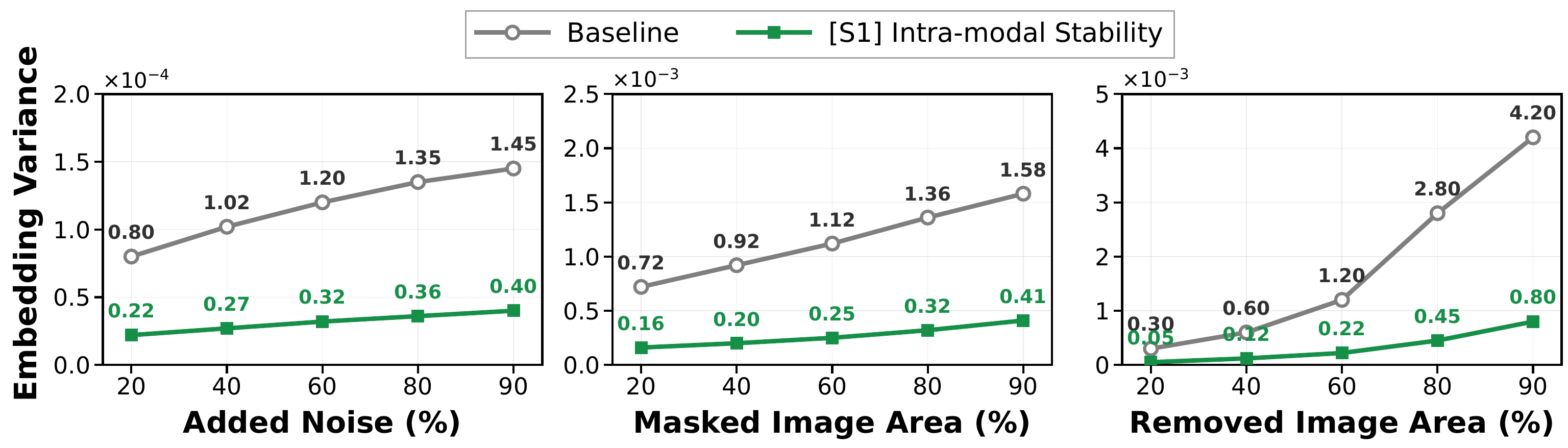}
        \caption{Embedding variance.}
        \label{fig:embedding_variance}
    \end{subfigure}
    \begin{subfigure}{\columnwidth}
        \centering
\includegraphics[width=\linewidth,keepaspectratio]{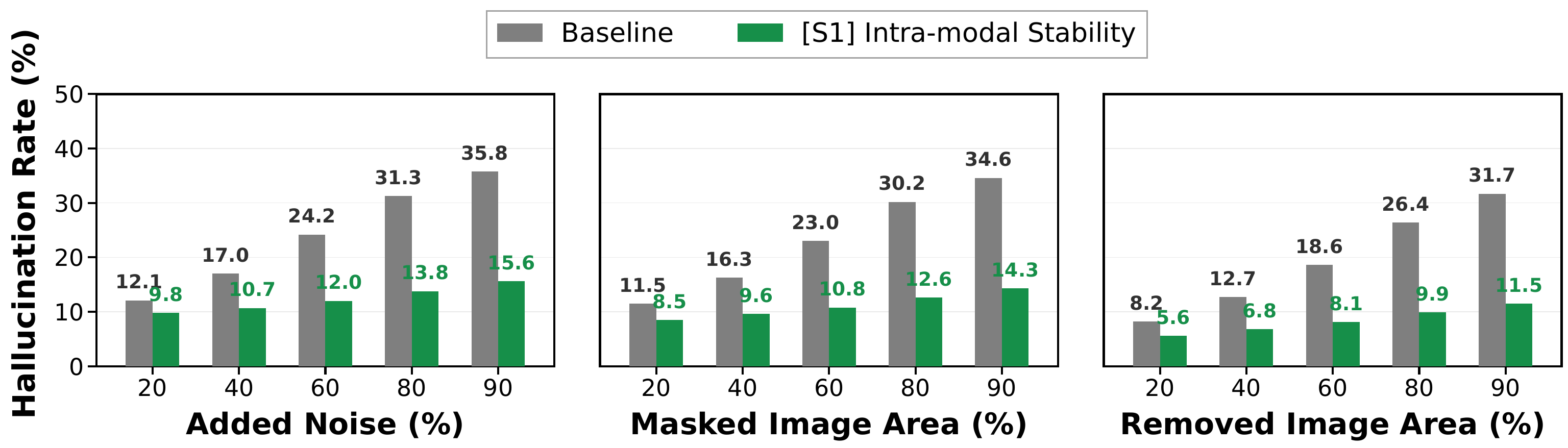}
        \caption{Hallucination rate.}
        \label{fig:hallucination_rate}
    \end{subfigure}
   \caption{\textbf{Sensitivity to image perturbations (LLaVA-1.6).}
\textbf{(a)} The \emph{baseline} model's embedding variance grows with
perturbation strength; training with our \textbf{[S1]} objective yields
lower variance.
\textbf{(b)} Hallucination rates evaluated on a fixed smaller AMBER subset
increase under the same perturbations, while \textbf{[S1]} consistently
reduces them. Baseline curves motivate the approach; \textbf{[S1]} curves
preview the effect of intervention, quantified in
Table~\ref{tab:ablation}.}
    \label{fig:sensitivity_analysis}
\end{figure}

\noindent\textbf{Feature instability and hallucination.}
Our starting point is an empirical observation about LVLM representations. Semantically equivalent observations of a scene should produce consistent embeddings. In practice, they do not: existing LVLMs produce substantially different visual features under controlled input perturbations such as additive noise, masking, or partial region removal. Figure \ref{fig:sensitivity_analysis} quantifies this on LLaVA-1.6. Embedding variance grows monotonically with perturbation strength (a), and hallucination rates grow under the same perturbations (b). Correlation alone does not establish causation. What we add is interventional evidence: when we directly train the representations to be perturbation-invariant (Stage 1 {\bf [S1]}, previewed in the same figure and ablated in Table~\ref{tab:ablation}), both embedding variance and hallucination rates drop substantially, before any cross-modal objective is applied. Taken together, the two observations support treating representation stability as an actionable training target, complementing the well-studied role of cross-modal misalignment \cite{jiang2024hallucination, ICLR2025_e1c73e95, ICLR2025_b73228d0}.
\vspace{2 mm}

\noindent\textbf{Implicit vs.\ explicit stabilization.}
Stability-motivated remedies exist, but current ones are \emph{explicit}: they impose stability from outside the trained model at inference time. Latent-space steering (VTI) \cite{liu2025reducing} shifts visual activations along pre-computed directions at every query. Contrastive and constrained decoding methods (VCD, OPERA) \cite{leng2024mitigating,huang2024opera} correct instability's downstream symptoms during generation. These interventions leave the underlying representation space unchanged and pay a cost on every query. We take the opposite route: build perturbation-invariance into the weights during fine-tuning, so the deployed model is stable without any intervention during inference. Training-time hallucination methods to date, such as supervised fine-tuning, RLHF, preference optimization, and contrastive alignment \cite{yu2024rlhf,rafailov2023direct,jiang2024hallucination, ICLR2025_e1c73e95, ICLR2025_b73228d0, yan2025task}, optimize output behavior or cross-modal correspondence; none optimizes intra-modal representation stability directly during training. Conversely, the methods that do target stability act only at inference time. 
\vspace{2 mm}

\noindent\textbf{Our approach.}
We propose \algo~(Implicit Feature Stabilization and Alignment), a two-stage training framework. Stage~1 (\textbf{[S1]}) stabilizes representations within each modality: visual embeddings are pulled toward perturbation-averaged anchors whose estimation error contracts at rate $1/\sqrt{K}$ in the number of views (Proposition~\ref{prop:anchor}), and generated-text embeddings are pulled toward human-corrected truthful-response anchors and repelled from hallucinated captions. Stage~2 (\textbf{[S2]}) then aligns the stabilized representations across modalities with bidirectional contrastive objectives plus a standard generation loss. The ordering is deliberate. Alignment acts on representations that are already low-variance, so positive pairs are matched on semantics rather than on appearance noise. Our contributions are:
\begin{enumerate}[leftmargin = 4 mm]
    \item We articulate the distinction between \emph{implicit} (training-time, weight-level) and \emph{explicit} (inference-time, intervention-level) feature stabilization for LVLMs, present correlational and interventional evidence that instability is an actionable factor in hallucination, and give a formal justification: a variance-contraction guarantee for perturbation-averaged anchors and a Lipschitz argument connecting embedding stability to stability of the output (Proposition \ref{prop:anchor}).
    \item We instantiate implicit stabilization in \algo, with fully specified anchor constructions (Algorithm \ref{alg:s1}), an asymmetric stop-gradient design related to but distinct from BYOL/SimSiam-style self-distillation \cite{grill2020bootstrap,chen2021exploring}, and a stabilize-then-align schedule.
    \item Across LLaVA-1.5-7B, LLaVA-1.6-7B, and Qwen3-VL-8B-Instruct, \algo\ reduces AMBER CHAIR by 46-63\% relative to the corresponding base models, improves ObjHal, MMHal, HallusionBench, and POPE, and preserves or improves VQA-v2 and TextVQA, at lower training cost than the strongest training-time baseline and with nothing extra running at inference.
\end{enumerate}

\raggedbottom
\section{Related Work}
\label{sec:related}
\noindent{\bf Hallucination in LVLMs.} Modern LVLMs~\cite{radford2021learning,alayrac2022flamingo,li2023blip,liu2023visual,liu2024improved,liu2024llavanext} remain vulnerable to hallucination despite strong benchmark performance; surveys \cite{huang2025survey,sahoo2024comprehensive,yin2024survey,chen2026survey} catalogue types and causes. Mitigation approaches \cite{zhu2026mitigating,agrawal2026towards,yu2026causally,fazli2026inject,yu2026optimizing,lim2026aligning,wu2025mitigating,ji2026causallens} divide into inference-time and training-time families.
\vspace{2 mm}

\noindent{\bf Explicit (inference-time) stabilization and correction.} Inference-time methods mitigate hallucinations without updating model parameters through decoding refinement, response verification, attention intervention, or post-hoc correction~\cite{park2025convis,huang2024opera,wu2025mitigating,wang2025mint,leng2024mitigating,yu2026causally,fazli2026inject,ji2026causallens}. Closest to our motivation is VTI~\cite{liu2025reducing}, which attributes hallucinations to unstable visual features and stabilizes them during inference. In contrast, \algo\ learns stable representations during training, eliminating the need for per-query intervention. On the same LLaVA-1.5 backbone, \algo\ outperforms VTI on adversarial POPE (Table \ref{tab:pope_inference_comparison}). More generally, inference-time methods improve generation without explicitly strengthening the underlying representations and often incur additional inference cost \cite{zhou2024revisiting,liu2025reducing}.
\vspace{2 mm}

\noindent{\bf Implicit (training-time) mitigation.} Training-based methods reduce hallucinations by improving supervision, preference alignment, or cross-modal learning. Representative approaches include contrastive learning and object-level supervision~\cite{jiang2024hallucination}, RLHF-based methods such as HALVA~\cite{ICLR2025_b73228d0}, RLHF-V~\cite{yu2024rlhf}, and Fact-RLHF~\cite{sun2024aligning}, preference optimization methods including DPO~\cite{rafailov2023direct,yang2025mitigating}, CHiP~\cite{ICLR2025_e1c73e95}, Antidote~\cite{wu2025antidote}, and recent preference learning approaches~\cite{lim2026aligning}, as well as textual embedding refinement~\cite{agrawal2026towards}. While these methods improve supervision or image--text alignment, they do not explicitly learn perturbation-invariant visual and textual representations, leaving feature instability largely unaddressed. In contrast, \algo\ first stabilizes visual and textual features before cross-modal alignment. Our method constructs perturbation-averaged visual anchors with lagged-parameter repulsion and ground-truth textual anchors with hallucinated-caption negatives, drawing inspiration from asymmetric self-supervised learning methods such as BYOL~\cite{grill2020bootstrap} and SimSiam~\cite{chen2021exploring}.

\begin{figure*}[t]
  \begin{center}
\includegraphics[width=\textwidth]{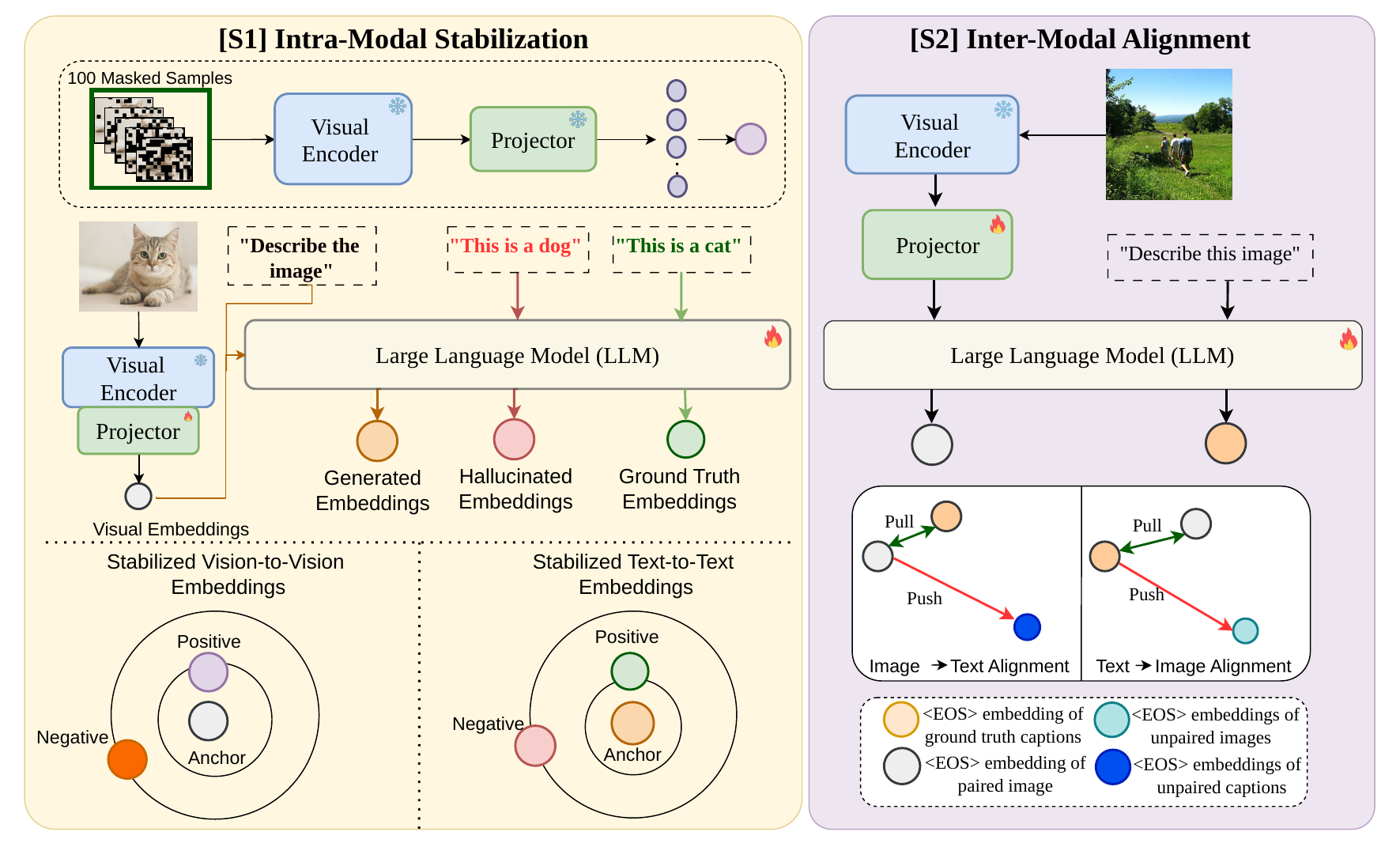}
    \caption{Overview of {\bf \algo}. {\bf [S1] Intra-modal stabilization} establishes perturbation-invariant anchors within each modality: perturbed views of each image are encoded, projected, and averaged into a low-variance visual anchor toward which the current embedding is pulled (and pushed from its lagged embedding); the ground-truth caption embedding anchors the textual branch, with hallucinated captions as negatives. {\bf [S2] Inter-modal alignment} aligns the stabilized representations bidirectionally (V$\rightarrow$T and T$\rightarrow$V), pulling paired image-text embeddings together and pushing unpaired combinations apart.}
    \label{fig:architecture}
  \end{center}
\end{figure*}

\section{The Proposed \algo\ Framework}

As illustrated in Figure \ref{fig:architecture}, \algo\ first learns stable representations within each modality (\textbf{[S1]}, {\em intra-modal stabilization}) and then aligns the stabilized representations across modalities (\textbf{[S2]}, {\em cross-modal alignment}), together with a standard generation objective that preserves generated language quality.
\vspace{2 mm}

\noindent\textbf{Problem setup.} Let $\{(v_i,t_i)\}_{i=1}^{N}$ be image-caption pairs. An LVLM comprises a frozen vision encoder $V_\theta$, a multimodal projector $F_\alpha$, and a language model $L_\beta$; following standard practice, only $(\alpha,\beta)$ are trained. Given $v_i$, the model generates $\hat{t}_i=L_\beta(F_\alpha(V_\theta(v_i)))$. For a text sequence $t$, we write $L_\beta(t)\in\mathbb{R}^d$ for the final-token hidden state of the decoder, used as a sentence-level embedding following \cite{liu2025reducing,jiang2024hallucination, ICLR2025_e1c73e95}.

\subsection{[S1]: Intra-Modal Stabilization}

\noindent\textbf{Vision-to-vision (V$\rightarrow$V) stabilization.}
A single image is one observation of a scene; it cannot tell us whether the representation of that scene is stable. Controlled perturbations provide multiple views of the same input while altering portions of its visual evidence. A well-grounded model should embed them consistently, and large dispersion among views indicates reliance on cues that the perturbations disturb. We therefore construct a low-variance \emph{anchor} by view averaging and train the representation toward it.

\begin{definition}[Perturbation-averaged visual anchor]
\label{def:anchor}
For image $v_i$, draw $K$ i.i.d.\ controlled perturbations $\{C_k(v_i)\}_{k=1}^{K}$ (random masking in our implementation), embed each view as $z_{ik}^v \coloneqq F_\alpha(V_\theta(C_k(v_i)))$, and define the anchor
$\bar z_i^{v} \coloneqq \frac{1}{K}\sum_{k=1}^{K} z_{ik}^v$,
treated as a constant (stop-gradient) in the loss.
\end{definition}

\begin{definition}[Lagged reference embedding]
\label{def:lagged}
At optimization step $s$, the \emph{lagged embedding} is the current image's embedding under the parameters of the previous step, $z_i^{v} \coloneqq \mathrm{sg}\!\left[F_{\alpha^{(s-1)}}(V_\theta(v_i))\right]$, where $\mathrm{sg}[\cdot]$ denotes stop-gradient. It represents the pre-update, potentially perturbation-sensitive state of the representation.
\end{definition}
\noindent
The current (learnable) embedding is $z_i^{v^{+}} \coloneqq F_{\alpha}(V_\theta(v_i))$, with gradients through the projector. The V$\rightarrow$V objective is
\begin{equation}
\label{eq:vv}
\mathcal{L}_{V\to V} \!  \coloneqq \!
-\frac{1}{N}\sum_{i=1}^{N} \! \log
\frac{\mathrm{e}^{{\mathrm{sim}(z_i^{v^+},\bar z_i^{v})}/{\tau_{vv}}}}
{\mathrm{e}^{{\mathrm{sim}(z_i^{v^+},\bar z_i^{v})}/{\tau_{vv}}} + \mathrm{e}^{{\mathrm{sim}(z_i^{v^+},z_i^{v})}/{\tau_{vv}}}},
\end{equation}
where $\mathrm{sim}(\cdot,\cdot)$ is cosine similarity and $\tau_{vv}$ a temperature.
\vspace{2 mm}

\noindent\textbf{Why repel from the lagged embedding?}
The repulsion term is an anti-collapse and escape mechanism, not a semantic judgment about the previous state. With the anchor held constant, the attraction term alone admits degenerate optima in which the projector barely moves. Repelling from the lagged embedding breaks this symmetry: each update must move measurably away from the pre-update state and toward the low-variance anchor. The construction belongs to the same family as the predictor and stop-gradient asymmetries that prevent collapse in BYOL~\cite{grill2020bootstrap} and SimSiam~\cite{chen2021exploring}, with an explicit repulsive term in place of their architectural asymmetry. Nor can the repulsion discard grounded semantics wholesale, since the attraction target $\bar z_i^v$ is built from views of the same image: the optimization moves within the image's semantic neighborhood, away only from its unstable directions.
\vspace{2 mm}

\noindent\textbf{Text-to-text (T$\rightarrow$T) stabilization.}
For each $v_i$ we use the human-corrected truthful response  $t_i$ and a hallucinated (rejected) caption $\tilde{t}_i$, with embeddings $z_i^{t} \coloneqq L_\beta(t_i)$ and $\tilde{z}_i^t\coloneqq L_\beta(\tilde{t}_i)$. Let $z_i^{t^+}\coloneqq L_\beta(\hat t_i)$ denote the embedding of the model's generated response $\hat t_i$; gradients flow through the decoder's hidden states on the generated token sequence (token identities are held fixed), so the objective shapes the representation of what the model actually generates. The generated embedding is the anchor of the contrastive comparison. It is pulled toward the ground-truth embedding and repelled from hallucinated-caption embeddings across the batch,
\begin{equation}
\label{eq:tt}
\mathcal{L}_{T \to T} \!
\coloneqq \! -\frac{1}{N} \! \sum_{i=1}^{N} \!
\log
\frac{
    \mathrm{e}^{{\operatorname{sim}(z_i^{t^+},\, z_i^t)}/{\tau_{tt}}}
}{\mathrm{e}^{{\operatorname{sim}(z_i^{t^+}\!, z_i^t)}/{\tau_{tt}}}  +  \sum_{j=1}^{N}\mathrm{e}^{{\operatorname{sim}(z_i^{t^+}\!, \tilde z_j^{t})}/{\tau_{tt}}} }.
\end{equation}
Here, $z_i^t$ acts as a fixed truthful anchor pulling the generated embedding toward grounded semantics, while hallucinated embeddings push it away from hallucination-prone regions.

\begin{algorithm}[t]
\caption{\algo\ Stage 1 (intra-modal stabilization)}
\label{alg:s1}
\begin{algorithmic}[1]
\Require pairs $\{(v_i,t_i,\tilde t_i)\}$, frozen $V_\theta$, projector $F_\alpha$, decoder $L_\beta$, views $K$, temperatures $\tau_{vv},\tau_{tt}$
\For{each optimization step $s=1,2,\dots$}
  \For{each $i$ in the batch}
    \State draw views $\{C_k(v_i)\}_{k=1}^K$;\; $\bar z_i^v \gets \frac1K\sum_k F_{\alpha^{(s)}}(V_\theta(C_k(v_i)))$ \Comment{stop-grad}
    \State $z_i^{v}\gets \mathrm{sg}[F_{\alpha^{(s-1)}}(V_\theta(v_i))]$;\quad $z_i^{v^+}\gets F_{\alpha^{(s)}}(V_\theta(v_i))$
   % \State $z_i^t \gets L_\beta(t_i)$;\; $\tilde z_i^t \gets L_\beta(\tilde t_i)$;\; $z_i^{t^+} \gets L_\beta(\hat t_i)$
   \State $z_i^t \gets \mathrm{sg}[L_\beta(t_i)]$;\;
$\tilde z_i^t \gets \mathrm{sg}[L_\beta(\tilde t_i)]$;\;
$z_i^{t^+} \gets L_\beta(\hat t_i)$
  \EndFor
  \State update $\alpha$ with $\nabla_\alpha \mathcal{L}_{V\to V}$ \big(Equation~\ref{eq:vv}\big);\; update $\beta$ with $\nabla_\beta \mathcal{L}_{T\to T}$ \big(Equation~\ref{eq:tt}\big)
\EndFor
\end{algorithmic}
\end{algorithm}
Overall, {\bf [S1]} optimizes
$\min_{\alpha} \mathcal{L}_{V \to V} + \min_{\beta} \mathcal{L}_{T \to T}$,
yielding stable embeddings in both modalities that serve as reliable references for {\bf [S2]}.

\subsection{Theoretical Justification}
\label{sec:theory}
Two questions deserve an answer: why is a perturbation-averaged anchor the right training target, and how does embedding stability transfer to the model's outputs?

\begin{proposition}[Anchor contraction and output stability]
\label{prop:anchor} Fix an image $v$ and let $z_k=F_\alpha(V_\theta(C_k(v)))$, $k=1,\dots,K$, be embeddings of i.i.d.\ semantics-preserving views with common mean $\mu_v=\mathbb{E}[z_1]$ (the perturbation-mean representation of $v$) and $\operatorname{Cov}[z_1]\preceq\sigma^2 I_d$. Then the anchor $\bar z=\frac1K\sum_k z_k$ satisfies
$$
\mathbb{E}\,\|\bar z-\mu_v\|_2^2 \;\le\; \frac{d\,\sigma^2}{K}.
$$
Suppose further the decoder's conditional output distribution is $L$-Lipschitz in the visual embedding under total variation, i.e., $\mathrm{TV}\!\left(p_\beta(\cdot\mid z),\,p_\beta(\cdot\mid z')\right)\le L\|z-z'\|_2$. Then for any set $\mathcal{H}$ of hallucinatory responses,
$$\left|\,\Pr_{p_\beta(\cdot\mid z)}[\mathcal{H}]-\Pr_{p_\beta(\cdot\mid \mu_v)}[\mathcal{H}]\,\right| \;\le\; L\,\|z-\mu_v\|_2 .
$$
\end{proposition}

\begin{proof}
The first bound is the variance of an i.i.d.\ mean:
$\mathbb{E}\|\bar z-\mu_v\|_2^2=\frac{1}{K^2}\sum_{k}\operatorname{tr}(\operatorname{Cov}[z_k])\le\frac{d\,\sigma^2}{K}$.
For the second,
$|\Pr_{p_\beta(\cdot\mid z)}[\mathcal H]-\Pr_{p_\beta(\cdot\mid\mu_v)}[\mathcal H]|
\le \mathrm{TV}(p_\beta(\cdot\mid z),p_\beta(\cdot\mid\mu_v))\le L\|z-\mu_v\|_2$,
by the definition of total variation and the Lipschitz assumption.
\end{proof}

\noindent
The first inequality says that the anchor's root-mean-square deviation
from the perturbation-mean representation $\mu_v$ contracts at rate
$1/\sqrt{K}$. Averaging views is therefore a statistically principled way to obtain a
lower-variance estimate of the perturbation-mean representation as $K$
increases. Empirically, performance generally improves with larger $K$, with diminishing returns at higher values, as reported in the supplementary material. The second inequality says that once embeddings concentrate near $\mu_v$, an input perturbation can change the probability of any hallucinatory output by at most $L$ times the embedding displacement it causes. The proposition should be read for what it establishes: a bound on how
much perturbations can shift the model's hallucination behavior, not a guarantee that hallucination rates fall. That they do fall under stabilization is an empirical result, shown by the ablations in
Table~\ref{tab:ablation}. As for the Lipschitz assumption, it is a mild smoothness condition on the decoder's conditional distribution over the continuous projected embedding; softmax decoders with bounded logit gradients satisfy it on any bounded embedding domain.

\subsection{[S2]: Cross-Modal Alignment}

Stable intra-modal representations are necessary but not sufficient. Reliable generation also requires accurate correspondence between modalities, since LVLMs connect pre-trained encoders and decoders independently through a light projector~\cite{liu2023visual,li2023blip,liu2024llavanext}. That architecture creates a representation gap; \algo\ closes it through training rather than architectural modification, and does so after stabilization, so that alignment matches semantics rather than appearance noise.

For each pair $(v_i,t_i)$, we take $u_i^{v}=L_\beta(F_\alpha(V_\theta(v_i)))$ and $u_i^{t}=L_\beta(t_i)$ (final decoder hidden states) and optimize a bidirectional InfoNCE objective in the style of CLIP~\cite{radford2021learning,jiang2024hallucination,leng2024mitigating}:
\begin{equation}
\label{eq:V-T}
  \mathcal L_{V \to T}
\coloneqq -\frac{1}{N}\sum_{i=1}^{N}
\log\frac{\mathrm{e}^{{\mathrm{sim}(u_i^v, u_i^t)}/{\tau_{vt}}}}
{\sum_{j=1}^{N}\mathrm{e}^{{\mathrm{sim}(u_i^v, u_j^t)}/{\tau_{vt}}}},
\end{equation}
\begin{equation}
\label{eq:T-V}
 \mathcal L_{T \to V}
\coloneqq -\frac{1}{N}\sum_{i=1}^{N}
\log\frac{\mathrm{e}^{{\mathrm{sim}(u_i^t, u_i^v)}/{\tau_{tv}}}}
{\sum_{j=1}^{N}\mathrm{e}^{{\mathrm{sim}(u_i^t, u_j^v)}/{\tau_{tv}}}}.
\end{equation}
{\bf [S2]} jointly optimizes $(\alpha,\beta)$ with
$\min_{\alpha,\beta}\,[\mathcal{L}_{V\rightarrow T}+\mathcal{L}_{T\rightarrow V}+\mathcal{L}_{G}]$,
where $\mathcal{L}_G$ is the standard supervised generation loss preserving fluency and downstream ability.
\vspace{2 mm}

\noindent\textbf{Distinct roles of Equation \eqref{eq:tt} and Equations \eqref{eq:V-T}-\eqref{eq:T-V}.}
Though contrastive in form, the two stages act on different objects with different negatives. Equation \eqref{eq:tt} is intra-textual: it shapes the embedding of the model's own generated text against truthful and hallucinated caption directions, with no visual term. Equations \eqref{eq:V-T}-\eqref{eq:T-V} are cross-modal, matching image embeddings to caption embeddings with unpaired-sample negatives. Figure~\ref{fig:s1_ablation} shows that the V$\rightarrow$V and T$\rightarrow$T stabilization branches are complementary, with their combination in [S1] yielding lower hallucination rates than either branch alone. Table~\ref{tab:ablation} further shows that adding [S2] after [S1] provides additional improvements, supporting the distinct and complementary roles of intra-modal stabilization and cross-modal alignment.
\vspace{2 mm}

\noindent\textbf{Implementation.}
We use $K=100$ masked views and temperature $\tau=0.07$ for all
contrastive objectives, and train {\bf [S1]} and {\bf[S2]} for 5 epochs each.
Sensitivity analyses over $\tau$ and $K$ are reported in
Table~\ref{tab:tau} and Table~\ref{tab:k_sensitivity}, respectively,
while Figure~\ref{fig:sensitivity_analysis} examines the effect of
varying perturbation strength.

\section{Experimental Evaluation}

We evaluate \algo\ on LLaVA-1.5-7B \cite{liu2024improved}, LLaVA-1.6-7B \cite{liu2024llavanext}, and Qwen3-VL-8B-Instruct \cite{DBLP:journals/corr/abs-2511-21631}, covering CLIP+Vicuna \cite{radford2021learning,zheng2023judging} and Qwen-family \cite{bai2023qwen,DBLP:journals/corr/abs-2511-21631} architectures.
\vspace{2 mm}

\noindent\textbf{Training data.}
Following prior preference-alignment work, we draw training data from RLHF-V \cite{yu2024rlhf}. Each selected item contains an image and a prompt with a preferred (chosen) response and a rejected response. We use the preferred response as the grounded textual target $t_i$ and the rejected response as the negative response $\tilde{t}_i$. \algo\ is trained on a 2K-sample subset.
\vspace{2 mm}

\noindent\textbf{Baselines and comparison protocol.} We separate two kinds of comparison and label them explicitly in Table \ref{tab:main_results}. \emph{Controlled comparisons} are methods we re-ran under our exact setting (same backbone, training data, and evaluation protocol): the base models and LLaVA-1.6-7B\,+\, DPO \cite{rafailov2023direct}, CHiP \cite{ICLR2025_e1c73e95} (LLaVA-1.6). \emph{Literature-reported results} (marked \litmark) are numbers taken from the original publications: RLHF-V \cite{yu2024rlhf} (Muffin~\cite{yu2023reformulating} architecture with BEiT-3 \cite{wang2023image} and Vicuna-13B), HALVA \cite{ICLR2025_b73228d0} (LLaVA-1.5),  which use their own training data and protocols. They contextualize our results but are not like-for-like, and we confine our strongest claims to the controlled block. We additionally compare against inference-time methods (Regular decoding, VCD \cite{leng2024mitigating}, OPERA \cite{huang2024opera}, VTI \cite{liu2025reducing}) on POPE.

\subsection{Results}
\noindent\textbf{Hallucination benchmarks.}
Table \ref{tab:main_results} reports results on ObjHal, MMHal, HallusionBench, and AMBER (generative). Across all three backbones, \algo\ consistently reduces hallucinations while maintaining competitive performance across complementary metrics. Compared with the corresponding base models, it improves all 11 metrics on both LLaVA-1.6 and Qwen3-VL, and all but one metric on LLaVA-1.5, where HallusionBench fA decreases slightly (24.9$\rightarrow$23.1). Under controlled comparisons, \algo\ outperforms DPO on 10 of the 11 metrics, with DPO retaining only a small advantage on HallusionBench fA (26.7 vs.\ 25.7). Compared with CHiP, evaluated under the same backbone, training data, and evaluation protocol, \algo\ achieves better performance on 9 of the 11 metrics, with CHiP outperforming \algo\ only on HallusionBench aA and AMBER Cog. The largest gains are observed on AMBER, where \algo\ reduces CHAIR from 7.8 to 4.2 (46\%) on LLaVA-1.5, from 8.3 to 3.1 (63\%) on LLaVA-1.6, and from 5.9 to 2.9 (51\%) on Qwen3-VL. Similar improvements are observed on ObjHal, where both response and mention rates are consistently reduced, and on MMHal, where \algo\ achieves the highest overall scores across all three backbones. Overall, these results demonstrate that the proposed training strategy generalizes effectively across diverse LVLM architectures.
\vspace{2 mm}

\noindent
\textbf{Fine-grained MMHal analysis.} Table~\ref{tab:mmhal_fine} reports the GPT-4-based fine-grained evaluation on MMHal-Bench. Compared with existing methods, ~\algo{} consistently improves overall performance while substantially reducing hallucinations. Specifically, INFUSE with LLaVA-1.6 achieves the highest overall GPT-4 score (3.87), whereas ~\algo{} with Qwen3-VL-8B-Instruct attains the lowest hallucination rate (15.12\%) and achieves the best performance in seven of the eight fine-grained evaluation categories. These consistent improvements across diverse reasoning types demonstrate that INFUSE enhances visual grounding and mitigates hallucinations in a broad range of multimodal reasoning scenarios.
\vspace{2 mm}

\noindent\textbf{Fine-grained discriminative analysis.}
Table~\ref{tab:amber_discriminative} reports AMBER's discriminative evaluation, including attribute-level (F1A) and relation-level (F1R) scores that probe finer-grained grounding than object existence. A natural worry about our design is that final-token anchors sacrifice fine-grained detail. The discriminative results say otherwise: for the backbones with corresponding base rows (LLaVA-1.6 and Qwen3-VL), \algo{} improves both F1A and F1R. In addition, on LLaVA-1.6, \algo{} improves F1R relative to the base model, whereas CHiP reduces it (62.0 vs.\ 69.6 for the base model). Token- and region-level extensions may capture still finer structure, which we discuss in the Limitations.
\vspace{2 mm}

\noindent\textbf{Comparison to explicit (inference-time) stabilization.} Table~\ref{tab:pope_inference_comparison} compares \algo{} with representative inference-time mitigation methods on the POPE adversarial split. On the shared LLaVA-1.5 backbone, \algo{} outperforms VTI, improving accuracy from 82.57\% to 83.56\% and F1 from 82.11 to 83.21, while also achieving higher precision and recall. With stronger backbones, \algo{} reaches 88.25\% accuracy with LLaVA-1.6 and 90.40\% with Qwen3-VL-8B, with the latter achieving the best overall F1 of 90.15.
\vspace{2 mm}

\noindent\textbf{General vision-language understanding.}
Table~\ref{tab:general_vl_performance} evaluates whether \algo\ preserves general multimodal capabilities on VQAv2~\cite{goyal2017making} and TextVQA~\cite{Singh_2019_CVPR}. Across all three backbones, \algo\ matches or improves the corresponding base models, achieving the best performance on LLaVA-1.6 and Qwen3-VL-8B. These results show that hallucination mitigation does not come at the expense of general vision-language understanding.
\vspace{2 mm}

\begin{table*}[!t]
\centering
\caption{Hallucination evaluation on four benchmarks. \litmark\ = literature-reported (original paper's backbone, data, and protocol; see text). Best and second-best are highlighted in \cellcolor{bestgreen}{green} and \cellcolor{secondorange}{orange}.}

\label{tab:main_results}
\setlength{\tabcolsep}{1.5pt}
\renewcommand{\arraystretch}{0.92}
\resizebox{0.98\linewidth}{!}{
\begin{tabular}{lccccccccccc}
\toprule
\multirow{2}{*}{\textbf{Model}}
& \multicolumn{2}{c}{\textbf{ObjHal}$^{\dagger}$}
& \multicolumn{2}{c}{\textbf{MMHal}$^{\ddagger}$}
& \multicolumn{3}{c}{\textbf{HallusionBench}$^{\S}$}
& \multicolumn{4}{c}{\textbf{AMBER}$^{\P}$}
\\
\cmidrule(lr){2-3}\cmidrule(lr){4-5}\cmidrule(lr){6-8}\cmidrule(lr){9-12}
& \textbf{R.}$\downarrow$ & \textbf{M.}$\downarrow$
& \textbf{Overall}$\uparrow$ & \textbf{R.}$\downarrow$
& \textbf{qA}$\uparrow$ & \textbf{fA}$\uparrow$ & \textbf{aA}$\uparrow$
& \textbf{CHAIR}$\downarrow$ & \textbf{Cover}$\uparrow$ & \textbf{Hal}$\downarrow$ & \textbf{Cog}$\downarrow$
\\
\midrule
\multicolumn{12}{l}{\emph{Base models (re-run)}}\\
LLaVA-1.5-7B~\cite{liu2024improved}
&46.3&22.6&2.4&52.1&10.6&24.9&46.9&7.8&51.0&36.4&4.2\\
LLaVA-1.6-7B~\cite{liu2024llavanext}
&14.1&7.4&2.8&42.7&15.8&20.8&51.6&8.3&61.0&48.6&4.2\\
Qwen3-VL-8B-Instruct~\cite{DBLP:journals/corr/abs-2511-21631}
&11.1&6.3&2.8&38.5&18.6&26.4&68.0&5.9&62.3&26.5&3.1\\
\midrule
\multicolumn{12}{l}{\emph{Literature-reported\litmark}}\\
RLHF-V\litmark~\cite{yu2024rlhf}
&12.2&7.5&2.5&51.0&--&--&--&6.3&46.1&25.1&2.1\\
HALVA\litmark~\cite{ICLR2025_b73228d0}
&--&--&--&--&13.9&20.1&49.1&--&--&--&--\\

\midrule
\multicolumn{12}{l}{\emph{Controlled comparison (re-run, same data and protocol)}}\\
LLaVA-1.6-7B + DPO
&14.0&7.1&2.3&46.2& 19.4&\cellcolor{secondorange}26.7&53.2&6.3&57.1&41.4&3.6\\
LLaVA-1.6-7B + CHiP~\cite{ICLR2025_e1c73e95}
&5.8 &3.9&3.1&40.2 &21.3&22.0&\cellcolor{secondorange}56.8 &3.9&55.4&25.7&\cellcolor{secondorange}1.8\\
\midrule
\multicolumn{12}{l}{\textbf{\algo\ (ours)}}\\
{\bf \algo} + LLaVA-1.5-7B
&5.1&3.7&3.78&18.75&39.1&23.1&49.7&4.2&\cellcolor{secondorange}61.1&28.2&2.1\\
{\bf \algo} + LLaVA-1.6-7B
&\cellcolor{secondorange}4.87&\cellcolor{secondorange}3.4&\cellcolor{bestgreen}\textbf{3.87}&\cellcolor{secondorange}16.02&\cellcolor{secondorange}41.0&25.7&56.3&\cellcolor{secondorange}3.1&\cellcolor{bestgreen}\textbf{68.1}&\cellcolor{secondorange}24.3&2.9\\
{\bf \algo} + Qwen3-VL-8B-Instruct
&\cellcolor{bestgreen}\textbf{4.13}&\cellcolor{bestgreen}3.2&\cellcolor{secondorange}3.82&\cellcolor{bestgreen}\textbf{15.12}&\cellcolor{bestgreen}\textbf{48.1}&\cellcolor{bestgreen}\textbf{32.0}&\cellcolor{bestgreen}\textbf{72.1}&\cellcolor{bestgreen}\textbf{2.9}&67.4&\cellcolor{bestgreen}\textbf{13.0}&\cellcolor{bestgreen}\textbf{1.0}\\
\bottomrule
\end{tabular}
}
$^{\dagger}$ ObjHal-COCO2017~\cite{lin2014microsoft,rohrbach2018object}.
$^{\ddagger}$ MMHal~\cite{sun2024aligning}.
$^{\S}$ HallusionBench~\cite{guan2024hallusionbench}.
$^{\P}$ AMBER~\cite{wang2023amber}.
\end{table*}

\begin{table*}[!t]
\centering
\caption{Fine-grained GPT-4 evaluation results on MMHal-Bench (MMHal)~\cite{sun2024aligning}.
Best and second-best results are highlighted in green and orange, respectively.
$\uparrow$ indicates higher is better, while $\downarrow$ indicates lower is better.}
\label{tab:mmhal_fine}

\renewcommand{\arraystretch}{0.98}
\setlength{\tabcolsep}{3.6pt}

\resizebox{\textwidth}{!}{%
\begin{tabular}{lcccccccccc}
\toprule

\multirow{2}{*}{\textbf{Model}}
& \multicolumn{10}{c}{\textbf{MMHal-Bench}} \\

\cmidrule(lr){2-11}

& \textbf{Overall}$\uparrow$
& \textbf{Hallu.}$\downarrow$
& \textbf{Attr.}$\uparrow$
& \textbf{Adv.}$\uparrow$
& \textbf{Comp.}$\uparrow$
& \textbf{Count.}$\uparrow$
& \textbf{Rel.}$\uparrow$
& \textbf{Env.}$\uparrow$
& \textbf{Hol.}$\uparrow$
& \textbf{Other}$\uparrow$ \\

\midrule

\multicolumn{11}{l}{\textit{Base models and existing methods}} \\[-1mm]

LLaVA~\cite{liu2024llavanext}
& 2.78 & 42.71 & 3.75 & 3.50 & 3.50
& 1.50 & 1.92 & 4.08 & 1.75 & 2.83 \\

LLaVA + DPO
& 2.3 & 46.2
& \cellcolor{orange!20}4.17
& 2.92 & 3.00 & 2.67 & 2.67
& \cellcolor{orange!20}4.25
& 1.17 & 1.00 \\

LLaVA + CHiP~\cite{ICLR2025_e1c73e95}
& 3.1 & 40.2
& \cellcolor{orange!20}4.17
& 3.33 & 2.67 & 2.67 & 2.25
& 4.08 & 1.67 & 1.92 \\

VTI~\cite{liu2025reducing}
& 2.90 & 51.00 & 3.75 & 2.42 & 3.50
& 2.42 & 2.83 & 3.50 & 2.42 & 2.33 \\

\midrule

\multicolumn{11}{l}{\textbf{INFUSE (ours)}} \\[-1mm]

INFUSE + LLaVA-1.5
& 3.78 & 18.75 & 3.67 & 3.58 & 3.50
& 2.83 & 2.58 & 4.07 & 1.71 & 2.83 \\

INFUSE + LLaVA-1.6
& \cellcolor{secondorange}{3.87}
& \cellcolor{secondorange}{16.02}
& 3.71
& \cellcolor{secondorange}{3.61}
& \cellcolor{secondorange}{3.52}
& \cellcolor{secondorange}{2.92}
& 2.57
& 4.12
& \cellcolor{secondorange}{3.42}
& \cellcolor{secondorange}{3.58} \\

INFUSE + Qwen3-VL-8B-Instruct
& \cellcolor{secondorange}{3.82}
& \cellcolor{bestgreen}\textbf{15.12}
& \cellcolor{bestgreen}\textbf{3.76}
& \cellcolor{bestgreen}\textbf{3.73}
& \cellcolor{bestgreen}\textbf{3.57}
& \cellcolor{bestgreen}\textbf{3.50}
& \cellcolor{bestgreen}\textbf{2.87}
& \cellcolor{bestgreen}\textbf{4.46}
& \cellcolor{bestgreen}\textbf{3.50}
& \cellcolor{bestgreen}\textbf{3.72} \\

\bottomrule
\end{tabular}%
}
\end{table*}

\begin{table*}[!t]
\centering
\vspace{1mm}

% Local caption formatting
\captionsetup{
    font=small,
    justification=centering,
    singlelinecheck=false,
    skip=0pt
}

% ============================================================
% TABLE 6: AMBER
% ============================================================
\begin{minipage}[t]{0.30\textwidth}
\vspace{0pt}
\centering

% Fixed caption area
\begin{minipage}[t][2.15cm][t]{\linewidth}
\captionof{table}{%
\textbf{Discriminative results on AMBER.}
Green and orange denote the best and second-best results.}
\label{tab:amber_discriminative}
\end{minipage}

\vspace{0.3mm}

\scriptsize
\renewcommand{\arraystretch}{1.12}
\setlength{\tabcolsep}{1.7pt}

\resizebox{\linewidth}{!}{%
\begin{tabular}{@{}lcccc@{}}
\toprule
\textbf{Model}
& \textbf{F1$\uparrow$}
& \textbf{F1E$\uparrow$}
& \textbf{F1A$\uparrow$}
& \textbf{F1R$\uparrow$} \\
\midrule

LLaVA-1.6 (7B)
& 87.0 & 95.1 & 81.5 & 69.6 \\

Qwen3-VL-8B
& 85.4 & 94.2 & 82.5 & 70.8 \\

LLaVA-1.6 + CHiP
& 86.9 & 98.3 & 80.3 & 62.0 \\

\midrule

\textbf{INFUSE} + LLaVA-1.5
& 87.1 & 95.5 & 81.6 & 70.1 \\

\textbf{INFUSE} + LLaVA-1.6
& \cellcolor{secondorange}{88.6}
& \cellcolor{bestgreen}\textbf{98.5}
& \cellcolor{secondorange}{82.1}
& \cellcolor{secondorange}{71.0} \\

\textbf{INFUSE} + Qwen3-VL-8B
& \cellcolor{bestgreen}\textbf{91.0}
& \cellcolor{secondorange}{97.8}
& \cellcolor{bestgreen}\textbf{84.0}
& \cellcolor{bestgreen}\textbf{78.0} \\

\bottomrule
\end{tabular}%
}

\end{minipage}
\hfill
% ============================================================
% TABLE 7: POPE
% ============================================================
\begin{minipage}[t]{0.40\textwidth}
\vspace{0pt}
\centering

% Fixed caption area
\begin{minipage}[t][2.15cm][t]{\linewidth}
\captionof{table}{%
\textbf{Comparison with inference-time mitigation methods on POPE.}
Green and orange denote the best and second-best results.}
\label{tab:pope_inference_comparison}
\end{minipage}

\vspace{0.2mm}

\scriptsize
\renewcommand{\arraystretch}{1.05}
\setlength{\tabcolsep}{2.0pt}

\resizebox{\linewidth}{!}{%
\begin{tabular}{@{}lcccc@{}}
\toprule
\textbf{Method}
& \textbf{Acc.}
& \textbf{Prec.}
& \textbf{Rec.}
& \textbf{F1} \\
\midrule

LLaVA-1.5 Regular\litmark
& 76.03 & 76.11 & 76.80 & 76.45 \\

LLaVA-1.5 + OPERA\litmark
& 80.88 & 82.16 & 79.76 & 80.94 \\

LLaVA-1.5 + VCD\litmark
& 77.31 & 73.43
& \cellcolor{secondorange}{86.47}
& 79.42 \\

LLaVA-1.5 + VTI\litmark
& 82.57 & 84.33 & 80.01 & 82.11 \\

\midrule

\textbf{INFUSE} + LLaVA-1.5
& 83.56 & 85.00 & 81.50 & 83.21 \\

\textbf{INFUSE} + LLaVA-1.6
& \cellcolor{secondorange}{88.25}
& \cellcolor{bestgreen}\textbf{92.69}
& 83.06
& \cellcolor{secondorange}{87.61} \\

\textbf{INFUSE} + Qwen3-VL-8B
& \cellcolor{bestgreen}\textbf{90.40}
& \cellcolor{secondorange}{92.55}
& \cellcolor{bestgreen}\textbf{87.88}
& \cellcolor{bestgreen}\textbf{90.15} \\

\bottomrule
\end{tabular}%
}

\end{minipage}
\hfill
% ============================================================
% TABLE 8: GENERAL VISION-LANGUAGE PERFORMANCE
% ============================================================
\begin{minipage}[t]{0.28\textwidth}
\vspace{0pt}
\centering

% Fixed caption area
\begin{minipage}[t][2.15cm][t]{\linewidth}
\captionof{table}{%
\textbf{General vision-language understanding.}
Green and orange denote the best and second-best results.}
\label{tab:general_vl_performance}
\end{minipage}

\vspace{0.3mm}

\scriptsize
\renewcommand{\arraystretch}{1.12}
\setlength{\tabcolsep}{1.8pt}

\resizebox{\linewidth}{!}{%
\begin{tabular}{@{}lcc@{}}
\toprule
\textbf{Model}
& \textbf{VQA-v2}
& \textbf{TextVQA} \\
\midrule

LLaVA-1.5 (Base)
& 76.2 & 56.4 \\

LLaVA-1.6 (Base)
& 80.2 & 60.9 \\

Qwen3-VL-8B (Base)
& 78.4 & 61.4 \\

\midrule

\textbf{INFUSE} + LLaVA-1.5
& 76.8 & 57.0 \\

\textbf{INFUSE} + LLaVA-1.6
& \cellcolor{bestgreen}\textbf{81.0}
& \cellcolor{secondorange}{61.5} \\

\textbf{INFUSE} + Qwen3-VL-8B
& \cellcolor{secondorange}{78.8}
& \cellcolor{bestgreen}\textbf{62.1} \\

\bottomrule
\end{tabular}%
}

\end{minipage}

\vspace{1mm}
\end{table*}

\begin{figure}[H]
    \centering
    \includegraphics[width=0.95\columnwidth]{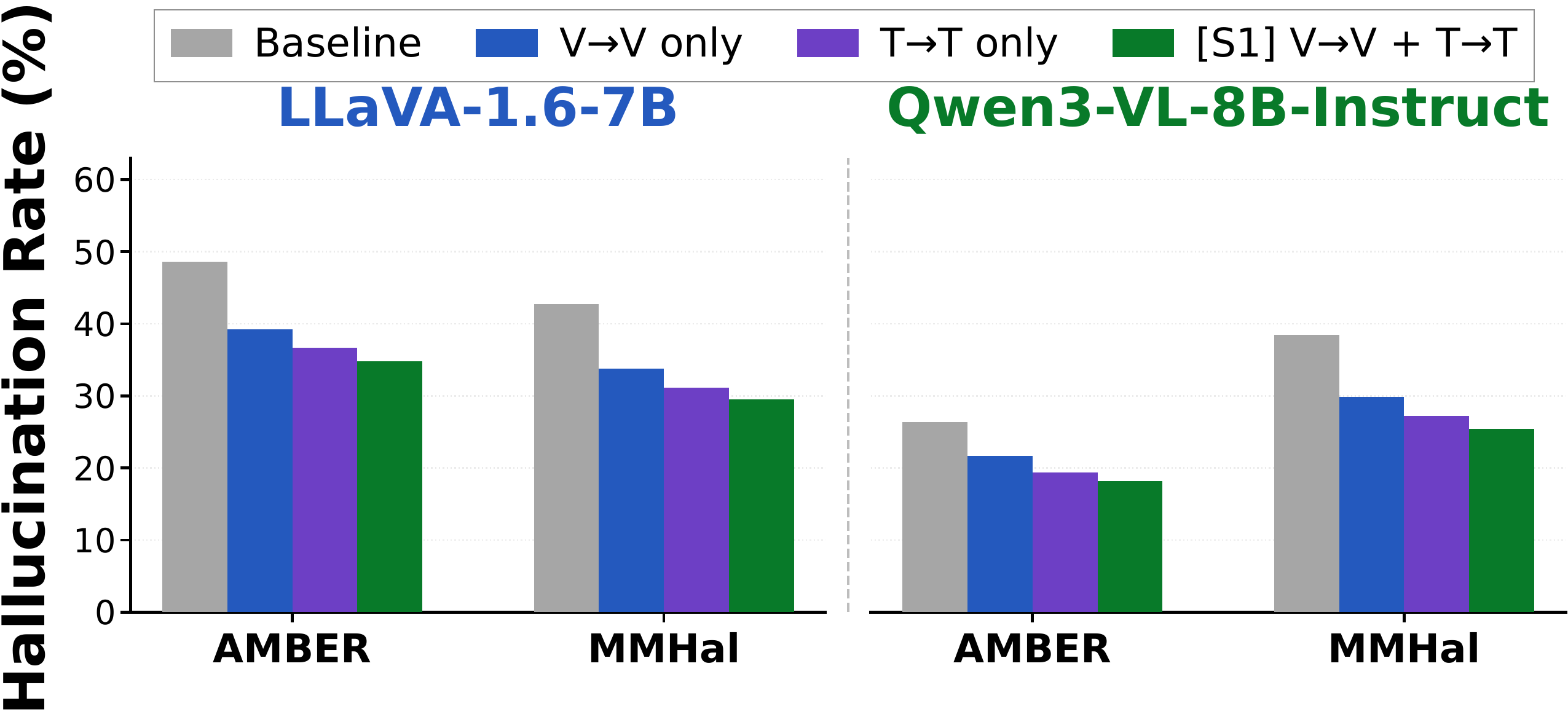}
    \caption{\textbf{{\bf [S1]} Ablation on hallucination rate (\%).} V$\rightarrow$V and T$\rightarrow$T each reduce hallucinations, while their combination achieves the lowest rates on AMBER and MMHal for LLaVA-1.6-7B and Qwen3-VL-8B-Instruct.}
    \label{fig:s1_ablation}
\end{figure}

\noindent
\textbf{Ablation study.} We evaluate the contribution of each component in \algo. Figure~\ref{fig:s1_ablation} shows that both V$\rightarrow$V and T$\rightarrow$T stabilization independently reduce hallucination rates on the AMBER discriminative benchmark and MMHal, while their combination ({\bf[S1]}) achieves the lowest hallucination rates. Table~\ref{tab:ablation} further demonstrates that these gains translate to improved generative performance on AMBER, with {\bf [S1]} consistently outperforming the corresponding baselines and {\bf [S2]} providing additional improvements.
\vspace{2 mm}

\noindent\textbf{Training cost.}
Table~\ref{tab:cost} reports aggregate GPU-hours, computed as the
number of GPUs multiplied by wall-clock training time. For this cost comparison, CHiP is evaluated using its official 5K+ training recipe, whereas \algo{} uses our 2K training set. CHiP requires 12.5 GPU-hours, compared with 6.2 GPU-hours for \algo{}. These values therefore compare the computational cost of the respective training recipes and should not be confused with the controlled comparison in Table~\ref{tab:main_results}, where CHiP is re-run using our matched
training-data setting.

\raggedbottom
\raggedbottom

\begin{table}[H]  
\centering
\caption{\textbf{Ablation study.}
{\bf [S1]}: $\mathcal{L}_{V\rightarrow V}+\mathcal{L}_{T\rightarrow T}$;
{\bf [S2]}: $\mathcal{L}_{V\rightarrow T}+\mathcal{L}_{T\rightarrow V}+\mathcal{L}_{G}$.
{\bf green} = best. $\uparrow$ higher is better, $\downarrow$ lower is better.}
\label{tab:ablation}
\resizebox{0.95\columnwidth}{!}{%
\begin{tabular}{lcccc}
\toprule
\textbf{Model} & \textbf{CHAIR}$\downarrow$ & \textbf{Cover}$\uparrow$ & \textbf{Hal}$\downarrow$ & \textbf{Cog}$\downarrow$ \\
\midrule
LLaVA-1.5-7B (base) & 7.8 & 51.0 & 36.4 & 4.2 \\
LLaVA-1.6-7B (base) & 8.3 & 61.0 & 48.6 & 4.2 \\
Qwen3-VL-8B-Instruct (base) & 5.9 & 62.3 & 26.5 & 3.1\\
LLaVA-1.6-7B + DPO  & 6.3 & 57.1 & 41.4 & 3.6 \\
LLaVA-1.6-7B + CHiP  & 3.9 & 55.4 & 25.7 & 1.8 \\
\midrule
\multicolumn{5}{l}{\textbf{\algo~(LLaVA-1.5-7B backbone)}} \\
 {\bf \algo}~{\bf [S1]}: $\mathcal{L}_{V\rightarrow V}+\mathcal{L}_{T\rightarrow T}$ & 7.3 & 52.3 & 32.5 & 3.7 \\
 {\bf \algo}~{\bf [S1] + \bf [S2]}: $\mathcal{L}_{V\rightarrow T}+\mathcal{L}_{T\rightarrow V}+\mathcal{L}_{G}$ &
4.2 & 61.1 & 28.2 & 2.1 \\
\multicolumn{5}{l}{\textbf{\algo~ (LLaVA-1.6-7B backbone)}} \\
 {\bf \algo}~{\bf [S1]}: $\mathcal{L}_{V\rightarrow V}+\mathcal{L}_{T\rightarrow T}$ & 6.8 & 60.2 & 35.4 & 3.6 \\
 {\bf \algo}~{\bf [S1] + \bf [S2]}: $\mathcal{L}_{V\rightarrow T}+\mathcal{L}_{T\rightarrow V}+\mathcal{L}_{G}$ &
\cellcolor{secondorange}{3.1} & \cellcolor{bestgreen}\textbf{68.1} & \cellcolor{secondorange}{24.3} & 2.9 \\
\multicolumn{5}{l}{\textbf{\algo~ ( Qwen3-VL-8B-Instruct backbone)}} \\
 {\bf \algo}~{\bf [S1]}: $\mathcal{L}_{V\rightarrow V}+\mathcal{L}_{T\rightarrow T}$ & 4.8 & 62.7 & 20.1 & \cellcolor{secondorange}{1.4} \\
 {\bf \algo}~{\bf [S1] + \bf [S2]}: $\mathcal{L}_{V\rightarrow T}+\mathcal{L}_{T\rightarrow V}+\mathcal{L}_{G}$ &
\cellcolor{bestgreen}\textbf{2.9} & \cellcolor{secondorange}{67.4} & \cellcolor{bestgreen}\textbf{13.0} & \cellcolor{bestgreen}\textbf{1.0} \\
\bottomrule
\end{tabular}%
}
{$^{\mathparagraph}$AMBER~\cite{wang2023amber}.}
\end{table}

\begin{table}[H]
\centering
\caption{\textbf{Training cost comparison} (LLaVA-1.6 backbone). CHiP cost is reported using its official training recipe, whereas \algo{} uses our 2K training setting. This cost comparison is therefore not the controlled 2K-data comparison reported in Table~\ref{tab:main_results}.}
\label{tab:cost}
\setlength{\tabcolsep}{4pt}
\begin{tabular}{lccc}
\toprule
\textbf{Method} & \textbf{Training Data} & \textbf{Epochs} & \textbf{GPU-hours} \\
\midrule
Base SFT & 5K & 5 & 12 \\
CHiP(official recipe) & 5K+ & 3 & 12.5 \\
\algo & 2K & 5+5 & \cellcolor{bestgreen}\textbf{6.2} \\
\bottomrule
\end{tabular}
\end{table}

\noindent{\textbf{Temperature sensitivity analysis.}} We evaluate the sensitivity of the contrastive temperature parameter $\tau$. As shown in Table~\ref{tab:tau}, ~\algo{} remains stable across a broad range of values ($0.01 \leq \tau \leq 0.20$), consistently reducing hallucinations relative to the baseline. The best performance is achieved at $\tau=0.07$, which we use throughout the paper.
\vspace{2 mm}

\begin{table}[H]
\centering
\caption{{\bf Sensitivity analysis} of temperature parameter $\tau$.}
\label{tab:tau}
\begin{tabular}{lcc}
\toprule
{\bf Temperature parameter}, $\tau$ & \textbf{CHAIR}$\downarrow$ & \textbf{Hal.}$\downarrow$ \\
\midrule
0.01 & 4.1 & 27.6 \\
0.05 & 3.5 & 25.8 \\
\textbf{0.07} (chosen) & \cellcolor{bestgreen}\textbf{3.1} & \cellcolor{bestgreen}\textbf{24.3} \\
0.10 & 3.3 & 25.1 \\
0.20 & 3.7 & 26.4 \\
\bottomrule
\end{tabular}
\end{table}

\begin{figure}[H]
  \centering
\includegraphics[width=0.95\textwidth]{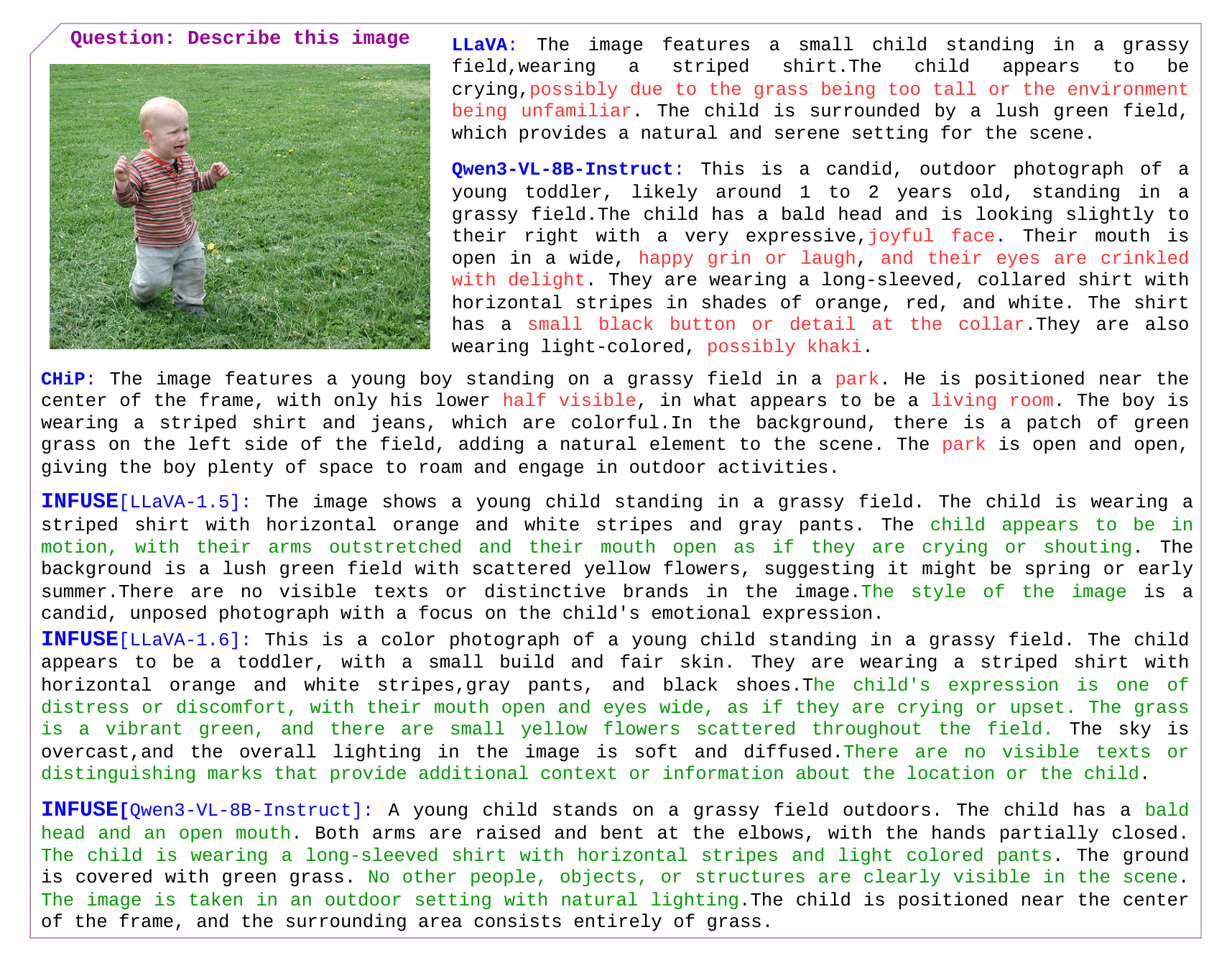}
  \caption{{\bf Qualitative comparison} on AMBER. LLaVA and CHiP produce hallucinated content: {\color{red}``grass being too tall''}, {\color{red}``living room''}, and {\color{red}``park''}; \algo\ ({\color{green}green}) describes visually supported content.}
  \label{fig:qual_baby}
\end{figure}

\noindent\textbf{Qualitative example.}
Figure~\ref{fig:qual_baby} shows a representative AMBER example. The baseline and CHiP outputs assert unsupported content (``grass being too tall,'' ``living room,'' ``park''); \algo\ sticks to the visible scene.
 
\section{Limitations}
\label{sec:limitations}
\algo\ operates on sentence-level (final-token) embeddings. The discriminative results suggest that attribute- and relation-level grounding improves anyway, but token- and region-level stabilization may capture finer detail and is a natural extension. Our textual negatives come from a single curated pool (RLHF-V rejected captions). The transfer we observe to AMBER, MMHal, HallusionBench, and POPE, none of which shares RLHF-V's distribution, suggests the learned stability is not tied to that pool's specific hallucination patterns, but broader negative sources deserve study. Our evaluation covers single-image, single-turn settings at the 7-8B scale; multi-image and multi-turn interactions, and larger models, remain open. Finally, Proposition~\ref{prop:anchor} controls perturbation-induced variation of the output distribution. Why stabilization also lowers absolute hallucination rates, which we establish empirically, is a question worth considering as a theory of its own.

\section{Conclusion}
We introduced the distinction between explicit (inference-time) and implicit (training-time) feature stabilization for LVLM hallucination mitigation, and instantiated the implicit route in \algo: perturbation-averaged and ground-truth anchors stabilize each modality, and bidirectional contrastive alignment then grounds the stabilized representations across modalities. The approach is theoretically motivated, with anchor contraction at rate $1/\sqrt K$ and Lipschitz control of output variation; cheap to train, at 6.2 GPU-hours (3.1 hours on two H100s) on 2K samples; and free at inference. Across three backbones, it reduces AMBER CHAIR by 46-63\% and AMBER hallucination rate by 23-51\% relative to the corresponding base models, while preserving VQA-v2 and TextVQA performance. We take this as evidence that implicit feature stabilization is a practical training principle for building more reliable LVLMs.

\newpage

\bibliographystyle{unsrtnat}
\bibliography{refs}

\clearpage

\section*{Appendix}

\setcounter{secnumdepth}{1}
\renewcommand{\thesection}{\Alph{section}}
\setcounter{section}{0}
\appendix
\noindent
This appendix is organized as follows.
Appendix~\ref{app:A} provides pseudocode and implementation-level details of the two training stages of \algo.
Appendix~\ref{app:B} describes training and data configurations, including architecture, optimization hyperparameters, and evaluation settings.
Appendix~\ref{app:C} evaluates robustness of the trained model under visual perturbations.
Appendix~\ref{app:D} reports additional quantitative results, including response-length statistics and fine-grained ObjHal metrics.
Appendix~\ref{app:E} presents representative hallucination cases for qualitative comparison against LLaVA, DPO, and CHiP.
 
\section{Algorithms for \algo}
\label{app:A}
 
Our method follows a two-stage training framework. {\bf [S1]} performs intra-modal stabilization, constructing perturbation-invariant anchors for both modalities; {\bf [S2]} performs inter-modal alignment on the stabilized representations. We describe both stages at the level of the implementation, expanding on Algorithm~1 of the main paper.
\vspace{2 mm}

\noindent\textbf{{\bf[S1]} Intra-modal stabilization.} We train on a subset $\mathcal{N}$ of the RLHF-V dataset~\cite{yu2024rlhf} with $|\mathcal{N}| = N = 2000$ samples. Each sample is a triplet $(v_i, t_i, \tilde{t}_i)$: input image, ground-truth caption, and hallucinated (rejected) caption.
\vspace{2 mm}
 
\noindent\textbf{Masked-view visual anchors.} At each optimization step, for each image $v_i$ in the batch we generate $K=100$ masked views $\{C_k(v_i)\}_{k=1}^K$ with patch-mask ratio $\rho = 0.97$, embed each view through the frozen CLIP vision encoder $V_\theta$ and the current projector, $z_{ik}^v \coloneqq F_{\alpha}\!\big(V_\theta(C_k(v_i))\big)$, and form the anchor
\begin{equation}
\bar{z}_i^v \coloneqq \frac{1}{K}\sum_{k=1}^K z_{ik}^v ,
\end{equation}
treated as a constant (stop-gradient) in the loss, as in Definition~\ref{def:anchor} of the main paper. Averaging suppresses high-variance, instability-driven directions while preserving consistent semantic structure (Proposition~\ref{prop:anchor}). The clean-image embedding $z_i^{v^+} \coloneqq F_{\alpha}\!\big(V_\theta(v_i)\big)$, which receives gradients, is pulled toward $\bar{z}_i^v$ and repelled from the lagged embedding $z_i^{v}$ computed under the previous step's parameters with stop-gradient (Definition~\ref{def:lagged} of the main paper), yielding the visual stability loss $\mathcal{L}_{V\to V}$ of Equation~\eqref{eq:vv}. Since the vision encoder is frozen, the $K$ view embeddings require forward passes only.
\vspace{2 mm}
 
\noindent\textbf{Textual anchors.} The ground-truth caption is encoded under the current decoder parameters with stop-gradient to form the truthful anchor $z_i^{t} \coloneqq \mathrm{sg}\big[L_{\beta}(t_i)\big]$, and the hallucinated caption gives the negative direction $\tilde{z}_i^{t} \coloneqq \mathrm{sg}\big[L_{\beta}(\tilde{t}_i)\big]$. The model's response $\hat t_i$ is decoded from $v_i$, and $z_i^{t^+} \coloneqq L_{\beta}(\hat t_i)$ denotes its embedding; gradients flow through the decoder's hidden states on the generated token sequence (token identities held fixed). The textual stability loss $\mathcal{L}_{T\to T}$ of Equation~\eqref{eq:tt} pulls $z_i^{t^+}$ toward the truthful anchor and pushes it away from hallucinated-caption embeddings $\tilde{z}_j^{t}$ across the batch, discouraging drift of the generated representation into hallucination-prone regions of the text space.
\vspace{2 mm}

\noindent\textbf{{\bf[S1]} objective.} The overall objective combines both terms, $\mathcal{L}_{\mathrm{[S1]}} \coloneqq \mathcal{L}_{V\to V} + \mathcal{L}_{T\to T}$, minimized with AdamW \cite{loshchilov2017decoupled} over the projector and decoder parameters $(\alpha,\beta)$; since $\mathcal{L}_{V\to V}$ depends only on $\alpha$ and $\mathcal{L}_{T\to T}$ only on $\beta$, this is equivalent to the per-modality updates of Algorithm~1 in the main paper. The stage yields stabilized components $F^{(1)}_\alpha$ and $L^{(1)}_\beta$; detailed steps are in Algorithm~\ref{alg:stage1}.
\vspace{2 mm}

\begin{algorithm}[t]
  \caption{{\bf[S1]} Intra-modal stabilization (implementation)}
  \label{alg:stage1}
  \begin{algorithmic}[1]
    \Require $\mathcal{N} \coloneqq \{(v_i, t_i, \tilde{t}_i)\}_{i=1}^{N}$, epochs $T_1$, initial $F_{\alpha}, L_{\beta}$, step size $\eta$
    \Ensure $F^{(1)}_\alpha, L^{(1)}_\beta$
    \State Split $\mathcal{N}$ into mini-batches $\{\mathcal{S}_s\}_{s=1}^{S}$ of size $N_m$
    \For{$e = 1$ \textbf{to} $T_1$} \Comment{epoch loop}
      \For{each mini-batch $\mathcal{S}_s$} \Comment{one optimization step}
        \For{each $i \in \mathcal{S}_s$}
          \State draw views $\{C_k(v_i)\}_{k=1}^K$;\; $\bar z_i^v \gets \frac1K\sum_k F_{\alpha}\!\big(V_\theta(C_k(v_i))\big)$ \Comment{stop-grad}
          \State $z_i^{v} \gets \mathrm{sg}\big[F_{\alpha^{\mathrm{prev}}}\big(V_\theta(v_i)\big)\big]$;\quad $z_i^{v^+} \gets F_{\alpha}\big(V_\theta(v_i)\big)$
          \State $\hat t_i \gets \mathrm{DECODE}(L_\beta; v_i)$;\quad $z_i^{t^+} \gets L_{\beta}(\hat t_i)$
          \State $z_i^{t} \gets \mathrm{sg}\big[L_{\beta}(t_i)\big]$;\quad $\tilde z_i^{t} \gets \mathrm{sg}\big[L_{\beta}(\tilde t_i)\big]$
        \EndFor
        \State Compute $\mathcal{L}_{V \to V}$ on $\mathcal{S}_s$ via Equation~\eqref{eq:vv};\; $\mathcal{L}_{T \to T}$ via Equation~\eqref{eq:tt}
        \State $\mathcal{L}_{\mathrm{[S1]}} \gets \mathcal{L}_{V \to V} + \mathcal{L}_{T \to T}$
        \State $\alpha^{\mathrm{prev}} \gets \alpha$;\quad $(\alpha, \beta) \gets \mathrm{AdamW}\big((\alpha, \beta), \nabla_{(\alpha,\beta)}\mathcal{L}_{\mathrm{[S1]}}, \eta\big)$
      \EndFor
    \EndFor
    \State \Return $F^{(1)}_\alpha \gets F_{\alpha}$,\; $L^{(1)}_\beta \gets L_{\beta}$
  \end{algorithmic}
\end{algorithm}
 
\begin{figure}[t]
\centering
\includegraphics[width=0.9\linewidth]{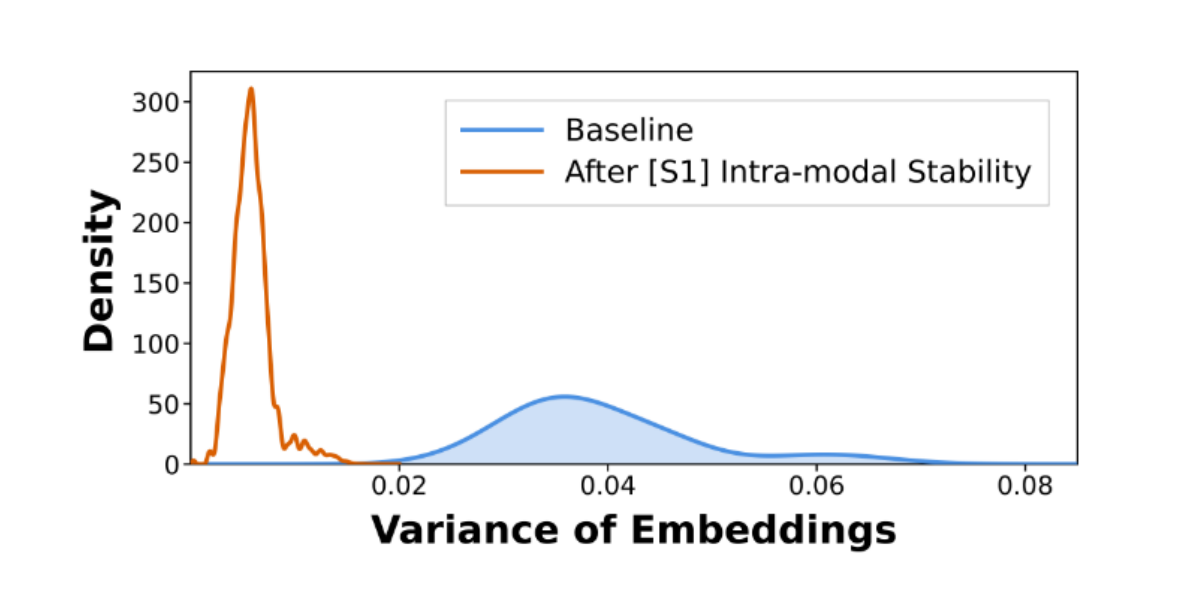}
\caption{\textbf{Embedding-variance distribution before and after intra-modal stabilization.} {\bf [S1]} shifts the entire distribution toward lower variance, indicating consistently more stable, perturbation-invariant visual representations across samples.}
\label{fig:density}
\end{figure}

\noindent\textbf{Embedding stability under perturbations.}
{\bf [S1]} explicitly reduces representational variance induced by input perturbations. As shown in Figure~\ref{fig:embedding_variance} of the main paper, the baseline exhibits progressively larger embedding variance as perturbation strength increases, whereas representations learned after {\bf [S1]} maintain substantially lower variance at all strengths. The density distribution in Figure~\ref{fig:density} corroborates this: {\bf [S1]} shifts the entire variance distribution toward lower values, so the improvement holds across perturbed samples rather than only on average. The stabilization objective achieves this by pulling the current representation toward the perturbation-averaged anchor while repelling the lagged embedding, reducing perturbation-induced drift without feature collapse.
\vspace{2 mm}
 
\noindent\textbf{{\bf[S2]} Inter-modal alignment.}
Starting from the stabilized projector and decoder of {\bf[S1]}, this stage enforces cross-modal grounding using the same $|\mathcal{N}| = N = 2000$ RLHF-V samples, now as image-caption pairs $(v_i, t_i)$. For each image, the visual-conditioned decoder embedding $u^v_i$ is obtained by passing $v_i$ through the frozen $V_\theta$, the projector $F_\alpha$, and the decoder $L_\beta$, taking the end-of-sequence (EOS) token embedding; the EOS embedding of the caption gives $u^t_i$. The bidirectional InfoNCE objectives of Equations~\eqref{eq:V-T}-\eqref{eq:T-V} pull each paired $(u^v_i, u^t_i)$ together and push unpaired combinations apart, and the supervised generation loss $\mathcal{L}_G$ preserves linguistic quality. The final objective, $\mathcal{L}_{\mathrm{[S2]}} = \mathcal{L}_{V\to T} + \mathcal{L}_{T\to V} + \mathcal{L}_G$, is minimized with AdamW over $(\alpha,\beta)$, yielding $F^{(2)}_\alpha$ and $L^{(2)}_\beta$. Detailed steps are in Algorithm~\ref{alg:stage2}.
 
\begin{algorithm}[t]
  \caption{{\bf[S2]} Inter-modal alignment (implementation)}
  \label{alg:stage2}
  \begin{algorithmic}[1]
    \Statex \textbf{Input:} $\mathcal{N} = \{(v_i, t_i)\}_{i=1}^N$, epochs $T_2$, step size $\eta$, $F^{(1)}_{\alpha}$, $L^{(1)}_{\beta}$
    \Statex \textbf{Output:} $F^{(2)}_\alpha$, $L^{(2)}_\beta$
    \State \textbf{Initialize:} $F_{\alpha} \gets F^{(1)}_{\alpha}$;\; $L_{\beta} \gets L^{(1)}_{\beta}$
    \State Split $\mathcal{N}$ into mini-batches $\{\mathcal{S}_s\}_{s=1}^{S}$ of size $N_m$
    \For{$e = 1$ \textbf{to} $T_2$}
      \For{$s = 1$ \textbf{to} $S$}
        \State $u^v_i \gets L_{\beta}\big(F_{\alpha}\big(V_\theta(v_i)\big)\big)$\;\; $\forall i \in \mathcal{S}_s$ \Comment{vision EOS}
        \State $u^t_i \gets L_{\beta}(t_i)$\;\; $\forall i \in \mathcal{S}_s$ \Comment{text EOS}
        \State Compute $\mathcal{L}_{V \to T}$ on $\mathcal{S}_s$ via Equation~\eqref{eq:V-T}
        \State Compute $\mathcal{L}_{T \to V}$ on $\mathcal{S}_s$ via Equation~\eqref{eq:T-V}
        \State Compute the generation loss $\mathcal{L}_{G}$ on $\mathcal{S}_s$
        \State $\mathcal{L}_{\mathrm{[S2]}} \gets \mathcal{L}_{V \to T} + \mathcal{L}_{T \to V} + \mathcal{L}_{G}$
        \State $(\alpha, \beta) \gets \mathrm{AdamW}\big((\alpha, \beta), \nabla_{(\alpha,\beta)}\mathcal{L}_{\mathrm{[S2]}}, \eta\big)$
      \EndFor
    \EndFor
    \State \Return $F^{(2)}_\alpha \gets F_{\alpha}$,\; $L^{(2)}_\beta \gets L_{\beta}$
  \end{algorithmic}
\end{algorithm}
 
\begin{figure}[t]
    \centering
    \includegraphics[width=0.9\columnwidth]{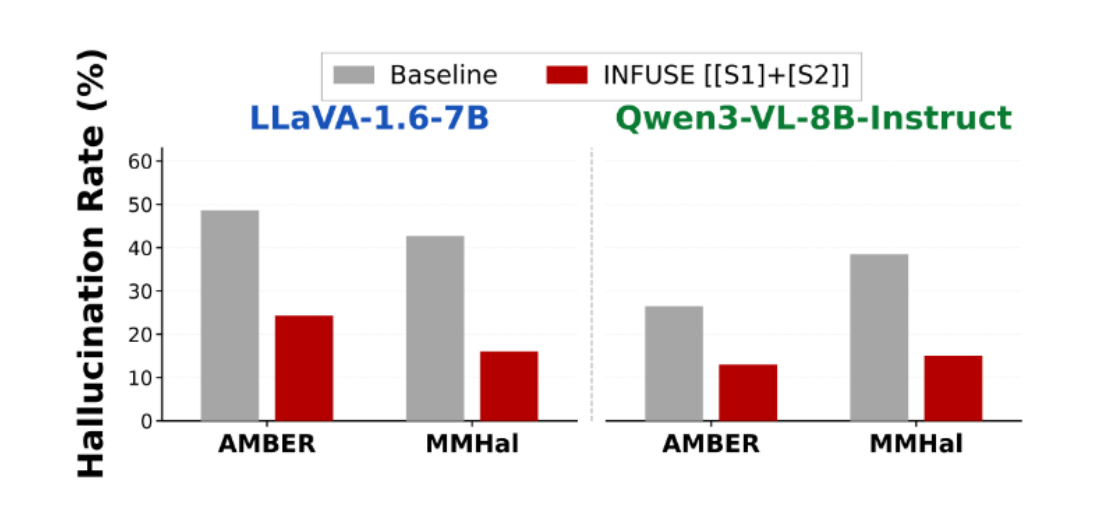}
    \caption{\textbf{Hallucination reduction after complete two-stage training.} Baseline models vs.\ the complete \algo\ framework ({\bf[S1]}+{\bf[S2]}) on AMBER and MMHal. Lower is better.}
    \label{fig:full_framework_hallucination}
\end{figure}

\noindent\textbf{Effect of the complete two-stage framework.}
Figure~\ref{fig:full_framework_hallucination} compares the base models with the final \algo\ model after both stages. Across LLaVA-1.6-7B and Qwen3-VL-8B-Instruct, the complete framework consistently reduces hallucination on AMBER and MMHal: for LLaVA-1.6-7B, from approximately $49\%$ to $24\%$ on AMBER and from $43\%$ to $16\%$ on MMHal; for Qwen3-VL-8B-Instruct, from approximately $27\%$ to $13\%$ and $39\%$ to $15\%$, respectively. The stable intra-modal representations learned in {\bf[S1]} thus provide a reliable foundation for the bidirectional alignment of {\bf[S2]}.
\vspace{2 mm}
 
\noindent\textbf{A note on training efficiency.} \algo\ is computationally lightweight for three reasons. First, it preserves the original architecture and keeps the vision encoder frozen, fine-tuning only the projector, LoRA ~\cite{hu2022lora} adapters on the last four decoder layers, and the associated parameters; this substantially lowers memory and optimization cost relative to full fine-tuning. Second, stability anchors are built from forward passes through the frozen encoder followed by simple averaging, with no auxiliary networks or learnable modules. Third, all stabilization and alignment objectives apply only during training; the inference pipeline is identical to the base model, with no added latency, memory, or deployment complexity.
 
\section{Training and Data Details}
\label{app:B}
 
\noindent\textbf{Training configuration for {\bf [S1]}.}
We train \algo\ on 2$\times$ NVIDIA H100 GPUs with bf16 mixed precision and fully synchronized gradient updates. For LLaVA-1.5-7B \cite{liu2024improved} and LLaVA-1.6-7B \cite{liu2024llavanext}, LoRA \cite{hu2022lora} adapters (rank 16, scaling 32) are attached only to the last four transformer layers of the Vicuna LLM; only these LoRA parameters and the vision--language projector are updated, while the CLIP ViT-L/336 vision encoder remains frozen. We optimize with AdamW \cite{loshchilov2017decoupled} at a peak learning rate of $2\times10^{-5}$ under cosine decay with no warm-up, batch size $N_m = 16$, for $T_1 = 5$ epochs on the $N = 2000$-sample RLHF-V subset. Each image is expanded into $K = 100$ masked views at a $97\%$ patch-mask ratio, and only the V$\to$V and T$\to$T objectives are active. Experiment tracking uses Weights \& Biases.
 \vspace{2 mm}
 
\noindent\textbf{Training configuration for {\bf [S2]}.}
{\bf [S2]} adds the cross-modal objectives (V$\to$T, T$\to$V) and the instruction-following generation loss. The trainable set is unchanged (last four decoder layers via LoRA and the projector); the vision encoder stays frozen to avoid re-introducing encoder-level instability during cross-modal alignment. We use AdamW with a base learning rate of $2\times10^{-5}$ and a cosine schedule, batch size $N_m=16$, for $T_2=5$ epochs. We use RLHF-V image-truthful-response pairs, where paired samples form the positives and other samples in the mini-batch provide unpaired negatives for the bidirectional contrastive objectives.
\vspace{2 mm}

\noindent\textbf{Qwen3-VL training configuration.}
For Qwen3-VL-8B-Instruct, we keep the vision encoder frozen and train the multimodal projector together with LoRA adapters on the last four language-model decoder layers. We use LoRA rank $r=16$, scaling factor $\alpha=32$, and dropout $0.05$. Both stages are trained using AdamW with a learning rate of $2\times10^{-5}$, weight decay $0.01$, BF16 precision, and an effective batch size of 16. Stage~{\bf [S1]} is trained for 5 epochs, and Stage~{\bf [S2]} is initialized from the Stage~{\bf [S1]} checkpoint and trained for another 5 epochs. We use $K=100$ masked views and a contrastive temperature of $\tau=0.07$ throughout training.
\vspace{2 mm}

\noindent\textbf{Sensitivity to the number of perturbation views.}
We evaluate the sensitivity of \algo{} to the number of perturbation views $K$ used to construct the perturbation-averaged visual anchor. Specifically, we vary $K\in\{10,25,50,100\}$ on LLaVA-1.6 while keeping all other training and evaluation settings fixed. As shown in Table \ref{tab:k_sensitivity}, increasing $K$ from 10 to 100 consistently improves hallucination mitigation, reducing CHAIR from 4.2 to 3.1 and the hallucination rate from 29.5\% to 24.3\%. Increasing $K$ further to 200 provides no additional improvement, with CHAIR and Hal.\ slightly increasing to 3.2 and 24.5\%, respectively. These results indicate diminishing returns at larger values of $K$ and motivate our choice of $K=100$ throughout the main
experiments.
\vspace{2 mm}

\begin{table}[t]
\centering
\caption{Sensitivity to the number of perturbation views $K$
on LLaVA-1.6. Lower values are better for both metrics.}
\label{tab:k_sensitivity}
\setlength{\tabcolsep}{8pt}
\renewcommand{\arraystretch}{1.08}
\begin{tabular}{ccc}
\toprule
\textbf{$K$} &
\textbf{CHAIR}$\downarrow$ &
\textbf{Hal.}$\downarrow$ \\
\midrule
10  & 4.2 & 29.5 \\
25  & 3.7 & 27.0 \\
50  & 3.3 & 25.3 \\
100 & \textbf{3.1} & \textbf{24.3} \\
\bottomrule
\end{tabular}
\end{table}

\noindent\textbf{Evaluation settings.}
We evaluate on five hallucination benchmarks: AMBER~\cite{wang2023amber}, Object HalBench~\cite{rohrbach2018object}, MMHal-Bench~\cite{sun2024aligning}, POPE~\cite{li2023evaluating}, and HallusionBench~\cite{guan2024hallusionbench}, summarized in Table~\ref{tab:eval_datasets}. Together they cover fine-grained object grounding, attribute faithfulness, multimodal reasoning consistency, and object-existence verification. All inference uses deterministic decoding (temperature $= 0$) for reproducibility.
\vspace{2 mm}

\noindent\textbf{Metrics.}
\textbf{Object HalBench} reports response-level and mention-level hallucination rates. \textbf{MMHal-Bench} reports GPT-4-assessed overall correctness and response-level hallucination rate. \textbf{HallusionBench} reports question-pair accuracy (qA), figure accuracy (fA), and overall accuracy (aA). \textbf{AMBER} (generative) reports CHAIR, object coverage (Cover), response-level hallucination rate (Hal), and cognition-driven hallucination rate (Cog); its discriminative track reports F1, F1E, F1A, and F1R. \textbf{POPE} reports accuracy, precision, recall, and F1 over Yes/No object-existence questions on MSCOCO images.

\begin{table}[t]
\centering
\setlength{\tabcolsep}{2pt}
\renewcommand{\arraystretch}{0.92}
\caption{\textbf{Evaluation datasets} used for hallucination analysis and multimodal grounding.}
\label{tab:eval_datasets}
\begin{tabularx}{\linewidth}{
@{}
>{\raggedright\arraybackslash}p{0.33\linewidth}
@{\hspace{4pt}}
>{\raggedright\arraybackslash}X
@{}
}
\toprule
\textbf{Dataset} & \textbf{Brief description} \\
\midrule
\textbf{Object HalBench}\\[-1pt]
{\footnotesize\citep{rohrbach2018object}}
&
Detects object hallucination in image captioning and open-ended descriptions.
\\
\textbf{MMHal-Bench}\\[-1pt]
{\footnotesize\citep{sun2024aligning}}
&
Multimodal hallucination dataset based on VQA, spanning multiple question types and object categories.
\\
\textbf{HallusionBench}\\[-1pt]
{\footnotesize\citep{guan2024hallusionbench}}
&
Robustness to visual illusions and prior-driven hallucinations across 346 images and 1,129 questions.
\\
\textbf{AMBER}\\[-1pt]
{\footnotesize\citep{wang2023amber}}
&
Evaluator-independent benchmark for generative hallucination and visual-textual consistency.
\\
\textbf{POPE}\\[-1pt]
{\footnotesize\citep{li2023evaluating}}
&
Binary object-existence benchmark with Yes/No questions over MSCOCO images.
\\
\bottomrule
\end{tabularx}
\end{table}
 
\section{Robustness Evaluation}
\label{app:C} 
To assess how well the \algo-trained model maintains performance under visual corruption, we evaluate several perturbation strategies applied to the input image at test time, with the model fixed. All rows use \algo\,+\,LLaVA-1.5.
\vspace{2 mm}
 
\noindent\textbf{Perturbation details.}
\textit{Masking (0.3 / 0.6)} removes $30\%$ or $60\%$ of spatial patches at random, simulating partial occlusion at two severities. \textit{Gaussian noise} adds pixel-level perturbations $\epsilon \sim \mathcal{N}(0, \sigma^2 I)$. \textit{Cropping} selects a random spatial window and resizes it to the original resolution, preserving global semantics while altering local structure. \textit{Rotation} applies a small random angular shift. Together, these span mild geometric variation through heavier distortion.
\vspace{2 mm}
 
\noindent\textbf{Results.}
Table \ref{tab:ours_rejection} summarizes results on HallusionBench (aA, qA) and AMBER (Cover, Hal). Clean images give the reference performance; corrupted inputs generally reduce accuracy and coverage. Rotation and cropping preserve performance best, while masking and Gaussian noise degrade semantic consistency more severely, with visible drops in qA and increased hallucination on AMBER. Perturbations that maintain global structure are therefore better tolerated, while heavy occlusion or random noise remains the harder regime.
 
\begin{table}[t]
  \centering
  \caption{Robustness of the \algo\,+\,LLaVA-1.5 model under test-time perturbations. aA and qA on HallusionBench; Cover and Hal on AMBER.}
  \label{tab:ours_rejection}
  \setlength{\tabcolsep}{3pt}
  \begin{tabular}{l|cc|cc}
    \toprule
    & \multicolumn{2}{c|}{\textbf{HallusionBench}} & \multicolumn{2}{c}{\textbf{AMBER}} \\
    \textbf{Perturbation}
            & aA$\uparrow$ & qA$\uparrow$ & Cover$\uparrow$ & Hal$\downarrow$ \\
    \midrule
    Clean images (reference) & \textbf{49.7} & \textbf{39.1} & \textbf{61.1} & \textbf{28.2} \\
    Masking (0.3)  & 45.3 & 23.7 & 49.2 & 30.2 \\
    Masking (0.6)  & 42.2 & 22.7 & 45.2 & 35.1 \\
    Gaussian       & 42.6 & 22.9 & 50.0 & 39.0 \\
    Cropping       & 45.0 & 36.1 & 47.0 & 31.2 \\
    Rotation       & 48.0 & 37.5 & 48.0 & 30.1 \\
    \bottomrule
  \end{tabular}
\end{table}
 
\section{Additional Quantitative Results}
\label{app:D}
 
\noindent\textbf{Response length across benchmarks.}
Response length is a useful behavioral signal: overly long outputs often accompany hallucination-prone generation (invented objects, fabricated attributes, scene embellishment), while overly short outputs can indicate lost informativeness. As shown in Table \ref{tab:avg_len}, \algo\ maintains output lengths comparable to DPO and CHiP and to the base model, indicating that its hallucination reductions do not come from truncating responses.
\vspace{2 mm}
 
\begin{table}[t]
\centering
\setlength{\tabcolsep}{3pt}
\renewcommand{\arraystretch}{0.95}
\caption{\textbf{Average output length (tokens).} Baseline rows use the LLaVA-1.6 backbone.}
\label{tab:avg_len}
\begin{tabularx}{\linewidth}{@{}Xccc@{}}
\toprule
\textbf{Model} & \textbf{AMBER} & \textbf{MMHal} & \textbf{ObjHal} \\
\midrule
LLaVA-1.6                     & 131 & 44 & 164 \\
LLaVA-1.6 + DPO               & 127 & 42 & 160 \\
LLaVA-1.6 + CHiP              & 124 & 43 & 151 \\
\textbf{\algo} + LLaVA-1.5    & 127 & 45 & 162 \\
\textbf{\algo} + LLaVA-1.6    & 125 & 45 & 164 \\
\textbf{\algo} + Qwen3-VL-8B-Instruct & 128 & 44 & 162 \\
\bottomrule
\end{tabularx}
\end{table}

\noindent\textbf{Fine-grained hallucination and grounding on ObjHal.}
Table~\ref{tab:objhal} decomposes ObjHal into response-level
hallucination (Resp. Hall), object-level hallucination (Obj. Hall),
response and object correctness, and object recall.
\algo{} + Qwen3-VL attains the lowest response-level hallucination
(4.13) and object-level hallucination (3.2), while CHiP achieves the
highest response correctness (95.08) and object correctness (96.79).
On LLaVA-1.5, response hallucination decreases from 46.3 to 5.1 and
object hallucination from 22.6 to 3.7. On LLaVA-1.6, the corresponding reductions are from 14.1 to 4.87 and from 7.4 to 3.4, respectively. These results show that \algo{} substantially reduces both response- and object-level hallucination while maintaining strong grounding performance.
 
\begin{table*}[t]
\centering
\caption{Fine-grained results on \textbf{Object HalBench (ObjHal)}. Bold = best, underline = second best. $\uparrow$ higher is better, $\downarrow$ lower is better. Baseline row uses the LLaVA-1.6 backbone.}
\label{tab:objhal}
\resizebox{\textwidth}{!}{
\begin{tabular}{lccccc}
\toprule
\textbf{Model} &
\textbf{Resp. Hall$\downarrow$} &
\textbf{Obj. Hall$\downarrow$} &
\textbf{Resp. Corr.$\uparrow$} &
\textbf{Obj. Corr.$\uparrow$} &
\textbf{Obj. Recall$\uparrow$} \\
\midrule
LLaVA-1.6 &
14.1 & 7.4 & 85.92 & 92.63 & \textbf{55.03} \\
LLaVA-1.6 + DPO &
14.0 & 7.1 & 88.97 & 93.39 & \underline{52.83} \\
LLaVA-1.6 + CHiP &
5.8 & \textbf{3.9} & \textbf{95.08} & \textbf{96.79} & 48.95 \\
\textbf{\algo} + LLaVA-1.5 &
5.1 & 3.7 & \underline{95.01} & 94.40 & 49.01 \\
\textbf{\algo} + LLaVA-1.6 &
\underline{4.87} & 3.4 & 94.86 & 95.40 & 51.08 \\
\textbf{\algo} + Qwen3-VL-8B-Instruct &
\textbf{4.13} & \underline{3.2} & 94.83 & \underline{96.34} & 52.75 \\
\bottomrule
\end{tabular}}
\end{table*}
 
\section{Representative Hallucination Examples}
\label{app:E}
 
\begin{figure*}[t]
    \centering
    \includegraphics[width=0.9\linewidth]{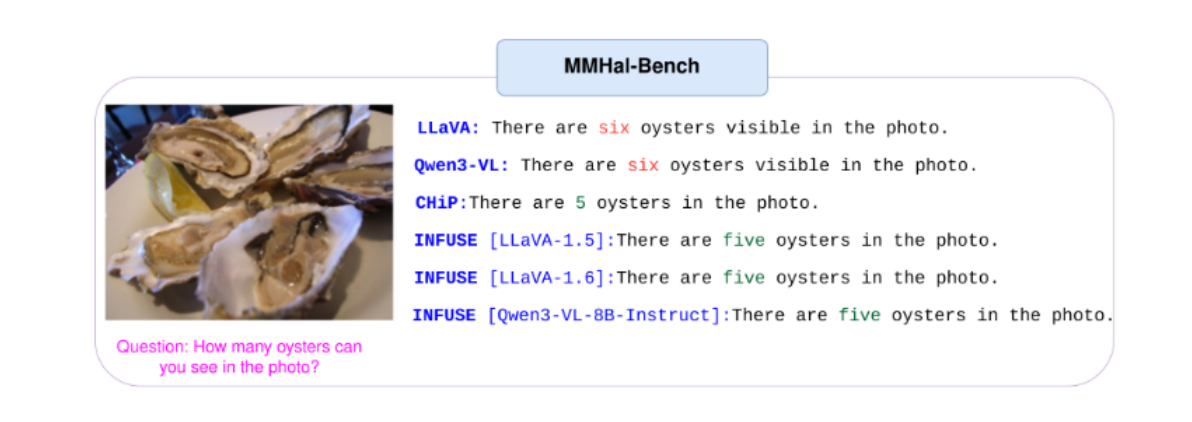}
    \caption{\textbf{MMHal-Bench: object-counting hallucination.} The base LLaVA and Qwen models hallucinate an extra object, claiming \textcolor{red}{six} oysters where five are visible. CHiP corrects the count, and \algo\ consistently identifies \textcolor{green}{five} oysters.}
    \label{fig:mmhal_oyster_reflection}
\end{figure*}

\begin{figure*}[t]
    \centering
    \includegraphics[width=0.8\linewidth]{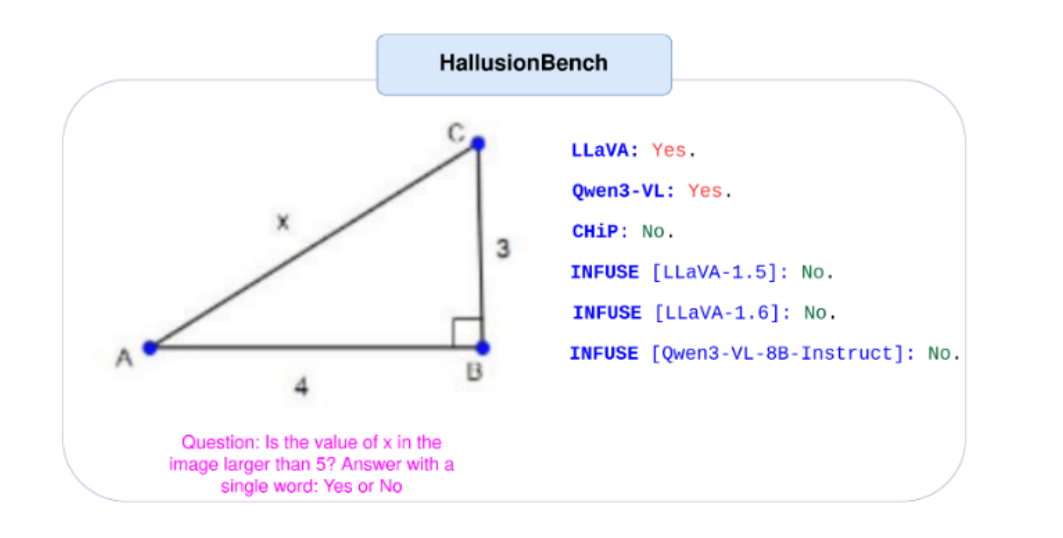}
    \caption{\textbf{HallusionBench: visual-geometric reasoning.} For a right triangle with legs 3 and 4, the hypotenuse is $x=5$. Asked whether $x$ exceeds 5, the base models incorrectly answer \textcolor{red}{Yes}; CHiP answers correctly, and \algo\ consistently answers \textcolor{green}{No}.}
    \label{fig:mmhal_pythagorean}
\end{figure*}

\begin{figure*}[t]
    \centering
    \includegraphics[width=0.8\linewidth]{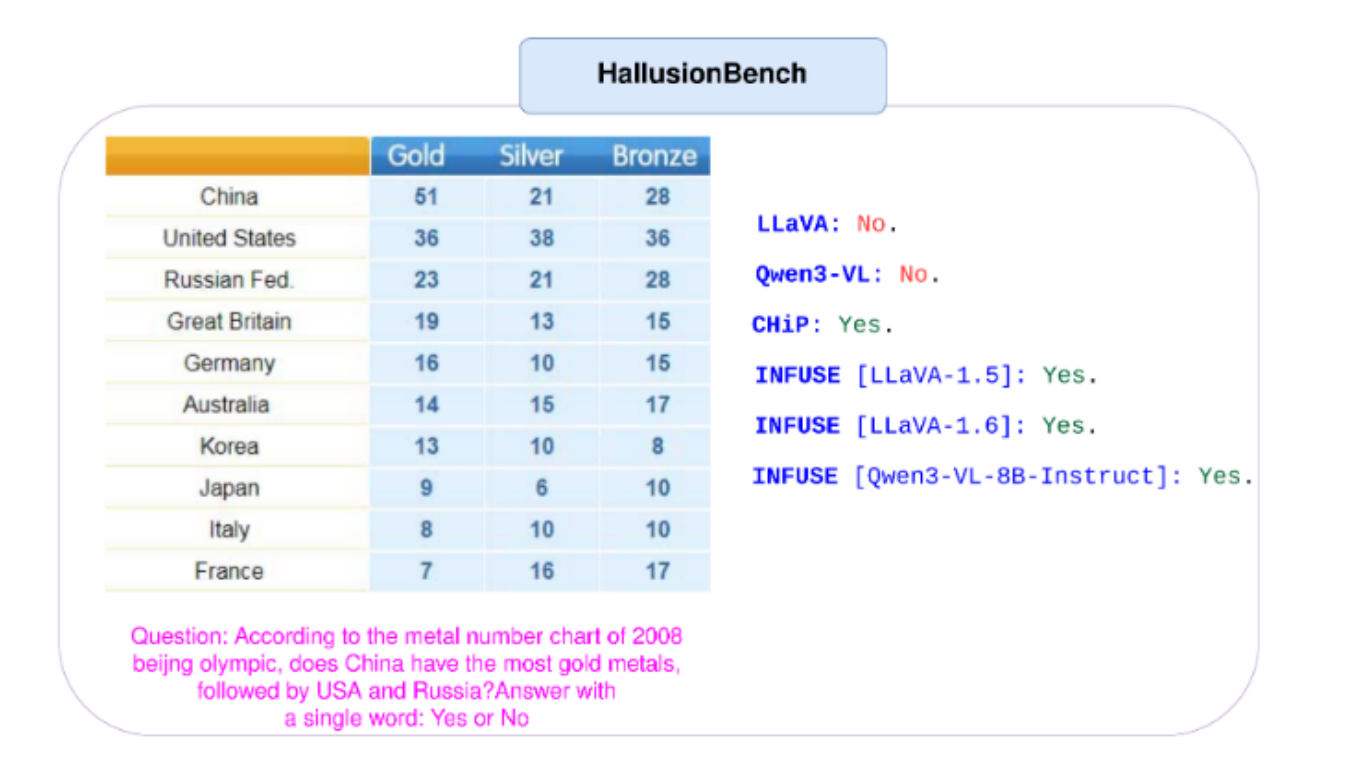}
    \caption{\textbf{HallusionBench (VS/table): table interpretation.} Given the 2008 Olympics medal table with China leading in golds, the base models incorrectly answer \textcolor{red}{No} when asked whether the shown ranking is correct; CHiP and \algo\ answer \textcolor{green}{Yes}, grounding in the visible entries.}
    \label{fig:hallu_olympic_table}
\end{figure*}

\begin{figure*}[t]
    \centering
    \includegraphics[width=0.8\linewidth]{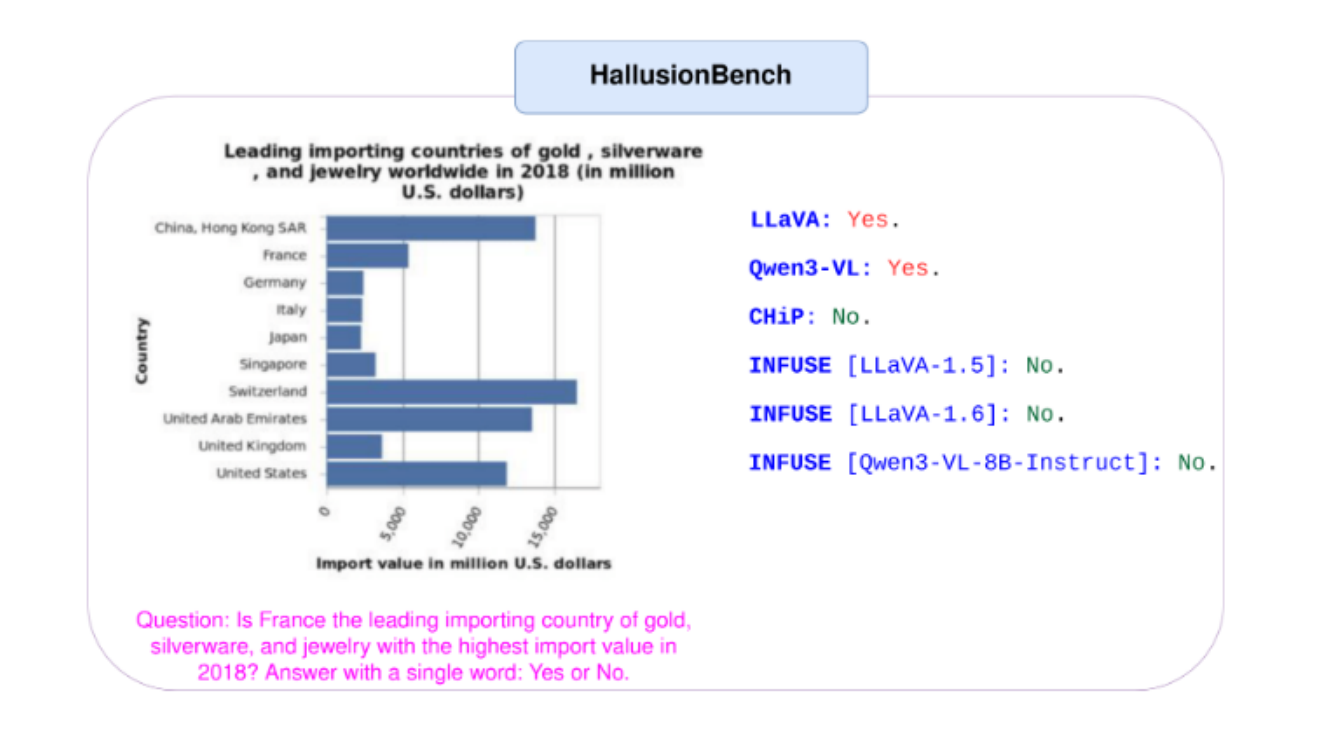}
    \caption{\textbf{HallusionBench (VS/chart): chart comprehension.} Switzerland has the longest bar in the import-value chart. Asked whether France leads, the base models incorrectly answer \textcolor{red}{Yes}, hallucinating a ranking that contradicts the bar lengths; CHiP and \algo\ answer \textcolor{green}{No}.}
    \label{fig:hallu_import_chart}
\end{figure*}

\begin{figure*}[t]
    \centering
    \includegraphics[width=0.9\linewidth]{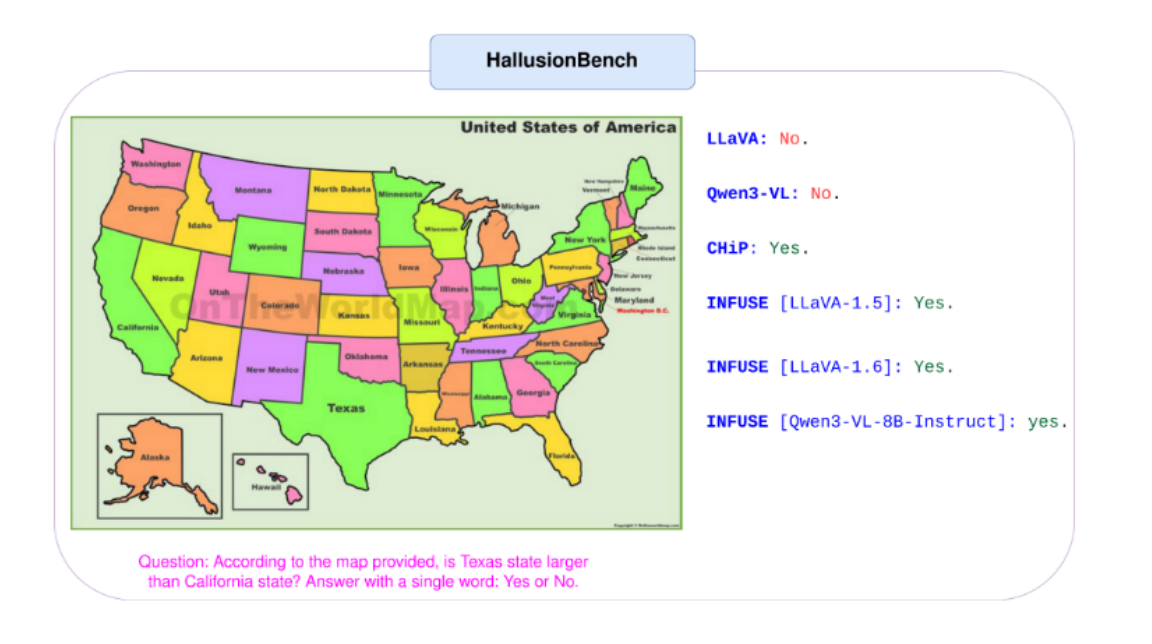}
    \caption{\textbf{HallusionBench (VS/map): geographic-size reasoning.} Texas is visibly larger than California, yet the base LLaVA model answers \textcolor{red}{No} when asked; CHiP and \algo\ answer \textcolor{green}{Yes}, matching the map.}
    \label{fig:hallu_map}
\end{figure*}

Figures~\ref{fig:mmhal_oyster_reflection}-\ref{fig:hallu_map} present five representative cases from MMHal-Bench and HallusionBench, covering object counting, geometric reasoning, table interpretation, chart comprehension, and map-based spatial comparison. Across these scenarios, the base models frequently hallucinate quantities, rankings, trends, or spatial relations unsupported by the visual evidence, while CHiP and especially \algo\ produce more conservative, visually grounded responses.
 
\end{document}